\documentclass[10pt,twocolumn,letterpaper]{article}

\usepackage{cvpr}              

\usepackage[utf8]{inputenc} 
\usepackage[T1]{fontenc}    
\usepackage{url}            
\usepackage{booktabs}       
\usepackage{amsfonts}       
\usepackage{nicefrac}       
\usepackage{microtype}      
\usepackage{color, colortbl}
\usepackage{array,multirow}
\usepackage{amsmath, mathtools, bm, bbm}
\usepackage{amsthm}
\usepackage{amssymb}
\usepackage{tabularx}
\usepackage{wrapfig}
\usepackage{xspace}
\usepackage{algorithm}
\usepackage{algpseudocode}
\usepackage[dvipsnames]{xcolor}
\usepackage[colorlinks,breaklinks=true,pagebackref]{hyperref}
\hypersetup{
    linkcolor=BrickRed,
    citecolor=Green
}

\newtheorem{prop}{Proposition}
\definecolor{lightgreen}{rgb}{0.5, 1.0, 0.5}

\newcommand{\norm}[1]{\left\lVert#1\right\rVert}

\newcommand{\support}{\mathbb{S}}
\newcommand{\query}{\mathbb{Q}}
\newcommand{\dbase}{\mathcal{D}_{\text{base}}}
\newcommand{\cseen}{\mathbb{C}_{\text{CS}}}
\newcommand{\cnovel}{\mathbb{C}_{\text{novel}}}
\newcommand{\cbase}{\mathbb{C}_{\text{base}}}
\newcommand{\cunseen}{\mathbb{C}_{\text{OS}}}

\newcommand{\inputspace}{\mathcal{X}}

\newcommand{\featurespace}{\mathcal{Z}}
\newcommand{\dphiteta}{\mathcal{D}_{\phi_{\theta}}}
\newcommand{\unitball}{\mathcal{B}_{2}}

\newcommand{\OSLO}{\textsc{OSLO}\xspace}
\newcommand{\knn}{$k$-\textsc{NN}}

\newcommand{\zbf}{\boldsymbol{z}}
\newcommand{\Zbf}{\boldsymbol{Z}}
\newcommand{\xibf}{\boldsymbol{\xi}}
\newcommand{\Xbf}{\boldsymbol{X}}
\newcommand{\xbf}{\boldsymbol{x}}

\newcommand{\ybf}{\boldsymbol{y}}

\newcommand{\pbf}{\boldsymbol{p}}
\newcommand{\mubf}{\boldsymbol{\mu}}
\newcommand{\Mubf}{\boldsymbol{\upsilon}}

\newcommand{\thetabf}{\boldsymbol{\theta}}

\definecolor{lightgray}{gray}{0.45}

\newcommand{\std}[1]{\color{lightgray}{\scriptsize{$\pm{#1}$}}}
\newcommand{\magicpar}[1]{\smallskip\noindent {\textbf{#1}}\enskip}

\usepackage[capitalize]{cleveref}
\crefname{section}{Sec.}{Secs.}
\Crefname{section}{Section}{Sections}
\Crefname{table}{Table}{Tables}
\crefname{table}{Tab.}{Tabs.}

\def\cvprPaperID{3866} 
\def\confName{CVPR}
\def\confYear{2023}

\begin{document}

\title{Open-Set Likelihood Maximization for Few-Shot Learning}

\author{%
  \textbf{Malik Boudiaf}\hspace{0.15em}$^{\ast}$\\
  \'ETS Montreal \\
  \and
  \textbf{Etienne Bennequin}\thanks{Equal contribution. Corresponding authors: \{malik.boudiaf.1@etsmtl.net, etienneb@sicara.com\}} \\
  Sicara -  MICS\hspace{0.04em}$^{\dagger}$
  \and
  \textbf{Myriam Tami} \\
  MICS\hspace{0.04em}$^{\dagger}$ \\
  \and
  \textbf{Antoine Toubhans} \\
  Sicara
  \and
  \textbf{Pablo Piantanida} \\
    ILLS - MILA - McGill - CNRS \\
  \and
  \textbf{Celine Hudelot} \\
  MICS\thanks{MICS, CentraleSupélec, Université Paris-Saclay}
  \and
  \textbf{Ismail Ben Ayed} \\
  \'ETS Montreal 
}

\maketitle

\begin{abstract}
    We tackle the Few-Shot Open-Set Recognition (FSOSR) problem, \ie classifying instances among a set of classes for which we only have a few labeled samples, while simultaneously detecting instances that do not belong to any known class. We explore the popular transductive setting, which leverages the unlabelled query instances at inference. Motivated by the observation that existing transductive methods perform poorly in open-set scenarios, we propose a generalization of the maximum likelihood principle, in which latent scores down-weighing the influence of potential outliers are introduced alongside the usual parametric model. 
    Our formulation embeds supervision constraints from the support set and additional penalties discouraging overconfident predictions on the query set. We proceed with a block-coordinate descent, with the latent scores and parametric model co-optimized alternately, thereby benefiting from each other. We call our resulting formulation \textit{Open-Set Likelihood Optimization} (\OSLO). \OSLO is interpretable and fully modular; it can be applied on top of any pre-trained model seamlessly. Through extensive experiments, we show that our method surpasses existing inductive and transductive methods on both aspects of open-set recognition, namely inlier classification and outlier detection.  Code is available at \url{https://github.com/ebennequin/few-shot-open-set}.
\end{abstract}

\section{Introduction}

    Few-shot classification consists in recognizing concepts for which we have only a handful of labeled examples. These form the \textit{support set}, which, together with a batch of unlabeled instances (the \textit{query set}), constitute a few-shot task. Most few-shot methods classify the unlabeled query samples of a given task based on their similarity to the support instances in some feature space \cite{snell2017prototypical}. This implicitly assumes a \textit{closed-set} setting for each task, i.e. query instances are supposed to be constrained to the set of classes explicitly defined by the support set.
    However, the real world is open and this closed-set assumption may not hold in practice, especially for limited support sets. Whether they are unexpected items circulating on an assembly line, a new dress not yet included in a marketplace's catalog, or a previously undiscovered species of fungi, \textit{open-set instances} occur everywhere. When they do, a closed-set classifier will falsely label them as the closest known class.
    
    This drove the research community toward open-set recognition \ie recognizing instances with the awareness that they may belong to unknown classes. In large-scale settings, the literature abounds of methods designed specifically to detect open-set instances while maintaining good accuracy on closed-set instances \cite{scheirer2012toward, bendale2016towards, zhou2021learning}. Very recently, the authors 
    of \cite{liu2020few} introduced a Few-Shot Open-Set Recognition (FSOSR) setting,
    in which query instances may not belong to any known class.
    The study in \cite{liu2020few}, together with other recent follow-up works \cite{jeong2021few, huang2022task}, exposed FSOSR to be a difficult task.


    To help alleviate the scarcity of labeled data, transduction \cite{vapnik2013nature}
    was recently explored for few-shot classification \cite{liu2018learning}, and has since
    become a prominent research direction, fueling a large body of works,
    e.g. \cite{veilleux2021realistic,dhillon2019baseline, liu2020prototype, ziko2020laplacian, boudiaf2020transductive,
    wang2020instance, hu2021leveraging, boudiaf2021few, martin2022towards}, among many others.
    By leveraging the statistics of the query set, transductive methods
    yield performances that are substantially better than their inductive counterparts
    \cite{boudiaf2020transductive,veilleux2021realistic} in the standard closed-set setting. 
    
    In this work, we seek to explore transduction for the FSOSR setting. We argue that
    theoretically, transduction has the potential to enable both classification
    and outlier detection (OD) modules to act symbiotically. 
    Indeed, the classification module can reveal valuable structure of the inlier's marginal distribution that the OD module seeks to estimate, such as the number of modes or conditional distributions, 
    while the OD part indicates the “usability”
    of each unlabelled sample. However, transductive principles currently adopted for few-shot learning heavily rely on
    the closed-set assumption in the unlabelled data, leading them to match
    the classification confidence for open-set instances with that of closed-set instances.
    In the presence of outliers, this not only harms their predictive performance
    on closed-set instances, but also makes prediction-based outlier detection substantially
    harder than with simple inductive baselines.
    

    \magicpar{Contributions.}
    In this work, we aim at designing a principled framework that reconciles transduction with the open nature of the FSOSR problem. Our idea is simple but powerful: instead of finding heuristics to assess the \textit{outlierness} of each unlabelled query sample, we treat this score as a latent variable of the problem. Based on this idea, we propose a generalization of the maximum likelihood principle, in which the introduced latent scores weigh potential outliers down, thereby preventing the parametric model from fitting those samples. Our generalization embeds additional supervision constraints from the support set and penalties discouraging overconfident predictions. We proceed with a block-coordinate descent optimization of our objective, with the closed-set soft assignments, \textit{outlierness} scores, and parametric models co-optimized alternately, thereby benefiting from each other. We call our resulting formulation \textit{Open-Set Likelihood Optimization} (\OSLO). \OSLO provides highly interpretable and closed-form solutions within each iteration for both the soft assignments, \textit{outlierness} variables, and the parametric model. Additionally, \OSLO is fully modular; it can be applied on top of any pre-trained model seamlessly.
    
    Empirically, we show that \OSLO significantly surpasses its inductive and transductive competitors alike for both outlier detection and closed-set prediction.  Applied on a wide variety of architectures and training strategies and without any re-optimization of its parameters, \OSLO's improvement over a strong baseline remains large and consistent. This modularity allows our method to fully benefit from the latest advances in standard image recognition. Before diving into the core content, let us summarize our contributions:
    \begin{enumerate}
        \item To the best of our knowledge, we realize the first study and benchmarking of transductive methods for the Few-Shot Open-Set Recognition setting. We reproduce and benchmark five state-of-the-art transductive methods.
        \item We introduce Open-Set Likelihood Optimization (\OSLO), a principled extension of the Maximum Likelihood framework that explicitly models and handles the presence of outliers. \OSLO is interpretable and modular \ie can be applied on top of any pre-trained model seamlessly.
        \item  Through extensive experiments spanning five datasets and a dozen of pre-trained models, we show that \OSLO consistently surpasses both inductive and existing transductive methods in detecting open-set instances while competing with the strongest transductive methods in classifying closed-set instances.
    \end{enumerate}

\section{Related Works}
    
\magicpar{Few-shot classification (FSC) methods.}
    Many FSC works involve episodic training \cite{Vinyals16}, in which a neural network acting as a feature extractor is trained on artificial tasks sampled from the training set. This replication of the inference scenario during training is intended to make the learned representation more robust to new classes. However, several recent works have shown that simple fine-tuning baselines are competitive in comparison to sophisticated episodic methods, e.g. \cite{Chen19, goldblum2020unraveling}, motivating a new direction of few-shot learning research towards the development of model-agnostic methods that do not involve any specific training strategy \cite{dhillon2019baseline}. 
    
\magicpar{Transductive FSC.}
    Transductive FSC methods leverage statistics of the query set as unlabeled data to improve performance, through model fine-tuning \cite{dhillon2019baseline}, Laplacian regularization \cite{ziko2020laplacian}, clustering \cite{lichtenstein2020tafssl}, mutual information maximization \cite{boudiaf2020transductive, veilleux2021realistic}, prototype rectification \cite{liu2020prototype}, or optimal transport \cite{bennequin2021bridging, hu2021leveraging, lazarou2021iterative}, among other transduction strategies.
    The idea of maximizing the likelihood of both support and query samples under a model parameterized by class prototypes is proposed by \cite{yang2020prototype} for few-shot segmentation. However, their method relies on the closed-set assumption. Differing from previous works, our framework leverages an additional latent variable, the \textit{inlierness} score.

\magicpar{Open-set recognition (OSR).}
    OSR aims to enable classifiers to detect instances from unknown classes \cite{scheirer2012toward}. Prior works address this problem in the large-scale setting by augmenting the SoftMax activation to account for the possibility of unseen classes \cite{bendale2016towards}, generating artificial outliers \cite{ge2017generative, neal2018open}, improving closed-set accuracy \cite{vaze2021open}, or using placeholders to anticipate novel classes' distributions with adaptive decision boundaries \cite{zhou2021learning}.
    All these methods involve the training of deep neural networks on a specific class set. Therefore, they are not fully fit for the few-shot setting. In this work, we use simple yet effective adaptations of OpenMax \cite{bendale2016towards} and PROSER \cite{zhou2021learning} as strong baselines for FSOSR.

\magicpar{Few-shot open-set recognition.}
    In the few-shot setting, methods must detect open-set instances while only a few closed-set instances are available.
    \cite{liu2020few} use meta-learning on pseudo-open-set tasks to train a model to maximize the classification entropy of open-set instances. \cite{jeong2021few} use transformation consistency to measure the divergence between a query image and the set of class prototypes. \cite{huang2022task} use an attention mechanism to generate a negative prototype for outliers. These methods require the optimization of a separate model with a specific episodic training strategy. 
    
    Nonetheless, as we show in \autoref{sec:experiments}, they bring marginal improvement over simple adaptations of standard OSR methods to the few-shot setting. In comparison, our method doesn't require any specific training and can be plugged into any feature extractor without further optimization.


\section{Few-Shot Open-Set Recognition} \label{sec:fsosr_setting}

    \magicpar{Model training.} Let us denote the raw image space $\inputspace$. As per the standard Few-Shot setting, we assume access to a $\textit{base}$ dataset ${\dbase = \{(\xbf_i,y_i)\}_{i=1...|\dbase|}}$ 
    with base classes $\cbase$, such that $\xbf_i \in \inputspace$ and $y_i \in \cbase$. We use $\dbase$ to train a feature extractor $\phi_{\thetabf}$. Our method developed later in \autoref{sec:open_set_likelihood}, freezes $\phi_{\thetabf}$ and performs inference directly on top of the extracted features for each task. \\
    
    \magicpar{Testing.} 
     Given a set of \textit{novel} classes $\cnovel$ disjoint from base classes \ie $\cnovel \cap \cbase = \emptyset$, a $K$-way FSOSR task is formed by sampling a set of $K$ \textit{closed-set} classes $\cseen \subset \cnovel$, a support set of labeled instances $\support = \{(\xbf_i,y_i) \in \inputspace \times \cseen \}_{i=1}^{|\support|}$ and a query set $\query = \{\xbf_i \in \inputspace \}_{i=|\support|+1}^{|\support|+|\query|}$. In the standard few-shot setting, the unknown ground-truth query labels $\{y_i\}_{i=|\support|+1}^{|\support|+|\query|}$ are assumed to be restricted to closed-set classes \ie $\forall i, ~y_i \in \cseen$. In FSOSR, however, query labels may also belong  to an additional set $\cunseen \subset \cnovel$ of \textit{open-set} classes \ie $\forall~i >  |\support|, ~y_i \in \cseen \cup \cunseen$ with $\cseen \cap \cunseen = \emptyset$. For easy referencing, we refer to query samples from the closed-set classes $\cseen$ as \textit{inliers} and to query samples from open-set classes $\cunseen$ as \textit{outliers}. For each query image $\xbf_i$, the goal of FSOSR is to simultaneously assign a closed-set prediction and an \textit{outlierness} (or equivalently \textit{inlierness}) score. \\

    \magicpar{Transductive FSOSR.} As a growing part of the Few-Shot literature, Transductive Few-Shot Learning assumes that unlabelled samples from the query set are observed at once, such that the structure of unlabelled data can be leveraged to help constrain ambiguous few-shot tasks. Transductive methods have achieved impressive improvements over inductive methods in standard closed-set FSC \cite{dhillon2019baseline, boudiaf2020transductive, ziko2020laplacian, hu2021leveraging}. We expect that transductive methods can help us improve overall open-set performance. While we find this to generally hold for closed-set predictive performance, we empirically show in \autoref{sec:experiments} that accuracy gains systematically come along significant outlier detection degradation, indicating that transductive methods are not equipped to handle open-set recognition. In the following, we take up the challenge of designing a transductive optimization framework that leverages the presence of outliers to improve its performance.


\section{Open-Set Likelihood} \label{sec:open_set_likelihood}

    \begin{figure}
        \centering
        \includegraphics[width=1\columnwidth]{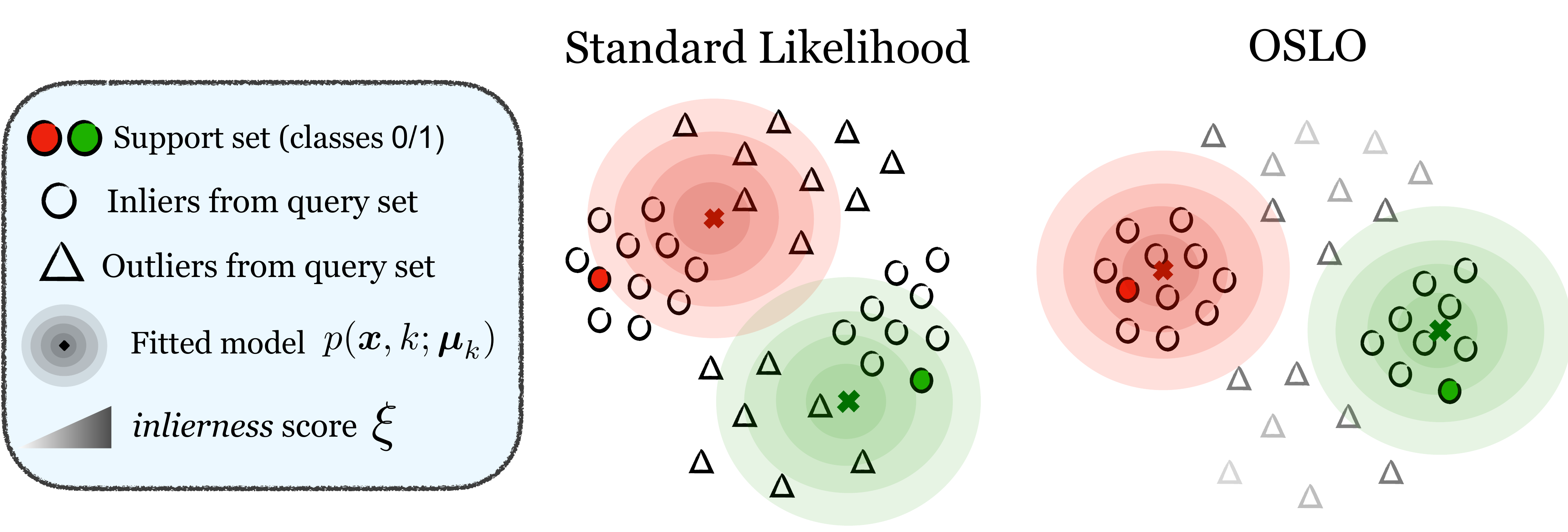}
        \caption{\textbf{Intuition behind \OSLO.} Standard transductive likelihood (left) tries to enforce high likelihood for all points, including outliers. \OSLO (right) instead treats the \textit{outlierness} of each sample as a latent variable to be solved alongside the parametric model. Besides yielding a principled \textit{outlierness} score for open-set detection, it also allows the fitted parametric model to effectively disregard samples deemed outliers, and therefore provide a better approximation of underlying class-conditional distributions.}
        \label{fig:likelihood}
    \end{figure}
    
    In this section, we introduce OSLO, a novel extension of the standard likelihood designed for transductive FSOSR. Unlike existing transductive methods, OLSO explicitly models and handles the potential presence of outliers, which allows it to outperform inductive baselines on both aspects of the open-set scenario. \\

    \magicpar{Observed variables.} We start by establishing the observed variables of the problem. As per the traditional setting, we observe images from the support set ${\{\xbf_i\}_{i=1}^{|\support|}}$ and their associated labels $\{y_i\}_{i=1}^{|\support|}$. The transductive setting also allows us to observe images from the query set. For notation convenience, we concatenate all images in $\Xbf={\{\xbf_i\}_{i=1}^{|\support|+|\query|}}$. \\
        
    \magicpar{Latent variables.} Our goal is to predict the class of each sample in the query set $\query$, as well as their \textit{inlierness}, i.e. the model's belief in a sample being an inlier or not. This naturally leads us to consider latent class assignments $\zbf_i \in \Delta^K$ describing the membership of sample $i$ to each closed-set class, with $\Delta^{K}=\{ \zbf \in [0, 1]^K: \zbf^T \mathbf{1}=1 \}$ the $K$-dimensional simplex. Additionally, we consider latent \textit{inlierness} scores $\xi_i \in [0, 1]$ close to 1 if the model considers sample $i$ as an inlier. For notation convenience, we consider latent assignments and \textit{inlierness} scores for all samples, including those from the support, and group everything in $\Zbf = \{\zbf_i\}_{i=1}^{|\support|+|\query|}$ and $\xibf = \{\xi_i\}_{i=1}^{|\support|+|\query|}$. Note that support samples are inliers, and we know their class. Therefore $\forall i \leq |\support|$, the constraints $\zbf_i=\ybf_i$ and $\xi_i=1$ will be imposed, where $\ybf_i$ is the one-hot encoded version of $y_i$  \\

    \magicpar{Parametric model.} The final ingredient we need to formulate is a parametric joint model over observed features and assignments. Following standard practice, we model the joint distribution as a balanced mixture of standard Gaussian distributions, parameterized by the centroids $\mubf=\{\mubf_1, \dots, \mubf_K\}$:
    \begin{align}
        p(\xbf, k; \mubf) = p(k) p(\xbf|k) \propto \exp(-\frac{\norm{\phi_{\thetabf}(\xbf)-\mubf_k}^2}{2})
    \end{align}
    As mentioned in \autoref{sec:fsosr_setting}, the feature extractor's parameters $\thetabf$ are kept frozen, and only $\mubf$ will be optimized. 

    \magicpar{Objective.} Using the i.i.d. assumption, we start by writing the standard likelihood objective:
    \begin{align}
        p(\Xbf, \Zbf; \mubf) = \prod_{i=1}^{|\support| + |\query|} \prod_{k=1}^K p(\xbf_i, k; \mubf)^{z_{ik}}
    \end{align}
    Without loss of generality, we consider the log-likelihood:
    \begin{align} \label{eq:standard_likelihood}
        \log(p(\Xbf, \Zbf; \mubf)) = \sum_{i=1}^{|\support| + |\query|} \sum_{k=1}^K z_{ik} \log(p(\xbf_i, k; \mubf))
    \end{align}
    Eq. \eqref{eq:standard_likelihood} tries to enforce a high likelihood of all samples under our parametric model $p$. This becomes sub-optimal in the presence of outliers, which should ideally be disregarded. \autoref{fig:likelihood} illustrates this phenomenon on a toy 2D drawing. To downplay this issue, we introduce \textit{Open-Set Likelihood Optimization} (\OSLO), a generalization of the standard likelihood framework, which leverages latent \textit{inlierness} scores to weigh samples:
    \begin{align} \label{eq:open_likelihood}
        \mathcal{L}_{\text{O}}(\Xbf, \Zbf, \xibf; \mubf) = \sum_{i=1}^{|\support| + |\query|} \xi_i \sum_{k=1}^K z_{ik} \log \left(p(\xbf_i, k; \mubf) \right)
    \end{align}
    Eq \eqref{eq:open_likelihood} can be interpreted as follows: samples believed to be inliers \ie $\xi_i \approx 1$ will be required to have high likelihood under our model $p$, whereas outliers won't. Note that $\xibf$ is treated as a variable of optimization, and is co-optimized alongside $\mubf$ and $\Zbf$. Finally, to prevent overconfident latent scores, we consider a \textit{penalty} term on both $\Zbf$ and $\xibf$:
    \begin{align}
        \mathcal{L}_{\text{soft}} =  \sum_{i=|\support|+1}^{|\support| + |\query|} \lambda_z \mathcal{H}(\zbf_i) + \lambda_\xi \mathcal{H}(\xibf_i)
    \end{align}
    where $\xibf_i = [1-\xi_i, \xi_i]$, and $\mathcal{H}(\pbf) = - \pbf^\top \log(\pbf)$ denotes the entropy, which encourages smoother assignments.

    \magicpar{Optimization.} We are now ready to formulate OSLO's optimization problem:
    \begin{align} \label{eq:oslo_objective}
            \max_{\mubf, \Zbf, \xibf} & \quad \mathcal{L}_{\text{O}}(\Zbf, \xibf, \mubf) + \mathcal{L}_{\text{soft}}(\Zbf, \xibf) \nonumber \\ 
            \text{s.t } & \quad \zbf_i \in \Delta^K, \quad \xi_i \in [0, 1] \quad \forall ~ i\\
            & \quad \zbf_i = \ybf_i, \quad \xi_i = 1,  \quad i \leq |\support|   \nonumber
    \end{align}
    Problem \eqref{eq:oslo_objective} is strictly convex with respect to each variable when the other variables are fixed.
    Therefore, we proceed with a block-coordinate ascent, which alternates three iterative steps, each corresponding to a closed-form 
    solution for one of the variables.
    \begin{prop} \label{prop:oslo}
        OSLO's optimization problem \eqref{eq:oslo_objective} can be minimized by alternating the following updates, with $\sigma$ denoting the sigmoid operation:
        \begin{align*}
            \xi_i ^{(t+1)} &=
                \begin{cases}
                    1 \quad \text{if } i \leq |\support| \\
                    \sigma \left(\displaystyle\frac{1}{\lambda_\xi} \displaystyle \sum_{k=1}^K z_{ik}^{(t)} \log p(\xbf_i,k; \mubf^{(t)}))\right) \text{else}
                \end{cases} \\
            \zbf_i^{(t+1)} &\propto 
                \begin{cases}
                    \ybf_i \quad \text{if } i \leq |\support| \\
                    \exp \left(\displaystyle\frac{\xi_i^{(t+1)}}{\lambda_z} \log p(\xbf_i, ~\cdot~; \mubf^{(t)}) \right)  \text{else}
                \end{cases} \\
            \mubf^{(t+1)}_k &= \frac{1}{\displaystyle\sum_{i=1}^{|\support|+ |\query|} \xi_i^{(t+1)} z_{ik}^{(t+1)}} \sum_{i=1}^{|\support| + |\query|} \xi_i^{(t+1)} z_{ik}^{(t+1)}\phi_{\theta}(\xbf_i)
        \end{align*}
    \end{prop}
    The proof of proposition \ref{prop:oslo} is performed by derivation of $\mathcal{L}_{\text{O}}(\Zbf, \xibf, \mubf) + \mathcal{L}_{\text{soft}}(\Zbf, \xibf)$ and deferred to the supplementary material. The optimal solution for the \textit{inlierness} score $\xi_i$ appears very intuitive and essentially conveys that samples with high likelihood under the current parametric model should be considered inliers. We emphasize that \textbf{beyond providing a principled \textit{outlierness score}, as $1-\xi_i$, the presence of $\xi_i$ allows to refine and improve the closed-set parametric model}. In particular, $\xi_i$ acts as a sample-wise temperature in the update of $\zbf_i$, encouraging outliers ($\xi_i \approx 0$) to have a uniform distribution over closed-set classes. Additionally, those samples contribute less to the update of closed-set prototypes $\mubf$, as each sample's contribution is weighted by $\xi_i$. 



\section{Experiments} \label{sec:experiments}

    \subsection{Experimental setup}
    
        \begin{table*}[t]
        \centering
        \small
        \caption{\textbf{Standard Benchmarking}. Evaluating different families of methods on the FSOSR problem on \textit{mini}-ImageNet and \textit{tiered}-ImageNet using a ResNet-12. For each column, a light-gray standard deviation is indicated, corresponding to the maximum deviation observed across methods for that metric. Best methods are shown in bold. Results marked with $^\star$ are reported from their original paper.}
        \resizebox{\textwidth}{!}{
            \begin{tabular}{lccccccccc}
                & & \multicolumn{8}{c}{\textbf{ \textit{mini}-ImageNet}} \vspace{0.5em}\\
                \toprule
                \multirow{3}{*}{Strategy} & \multirow{3}{*}{Method} & \multicolumn{4}{c}{1-shot} & \multicolumn{4}{c}{5-shot} \\
                \cmidrule(lr){3-6} \cmidrule(lr){7-10}
                & & Acc & AUROC & AUPR & Prec@0.9 & Acc & AUROC & AUPR & Prec@0.9 \\
                & & \std{0.72} & \std{0.79} & \std{0.69} & \std{0.47} & \std{0.44} & \std{0.73} & \std{0.61} & \std{0.56} \\
                \midrule
                \multirow{6}{*}{OOD detection} & \knn \cite{knn_detector} &      -     &         70.86 &       70.43 &             58.23 &      -     &         76.22 &       76.36 &             61.48  \\
                                                & IForest \cite{iforest_detector}  &      -     &         55.59 &       55.24 &             52.18 &      -     &         62.80 &       61.62 &             54.77  \\
                                                & OCVSM \cite{ocsvm_detector}  &      -     &         69.67 &       69.71 &             57.35 &      -     &         68.49 &       65.60 &             59.24  \\
                                                & PCA \cite{pca_detector}  &      -     &         67.23 &       66.50 &             56.67 &      -     &         75.24 &       75.53 &             60.73  \\
                                                & COPOD \cite{copod_detector}  &      -     &         50.60 &       51.85 &             50.92 &      -     &         51.63 &       52.65 &             51.31  \\
                                                & HBOS                 &      -     &         58.26 &       57.41 &             53.06 &      -     &         61.11 &       60.18 &             54.30  \\

                \toprule
                \multirow{3}{*}{Inductive classifiers} & SimpleShot \cite{wang2019simpleshot} &      65.90 &         64.99 &       63.78 &             55.77 &      81.72 &         70.61 &       70.06 &             57.91 \\
                                                        & Baseline ++ \cite{Chen19}&      65.81 &         65.15 &       63.85 &             55.87 &      81.86 &         66.37 &       65.58 &             56.33 \\
                                                        & FEAT  \cite{ye2020few}  &      67.23 &         52.45 &       54.44 &             50.00 &      82.00 &         53.25 &       56.48 &             50.00 \\

                \midrule
                \multirow{4}{*}{Inductive Open-Set} & PEELER$^\star$ \cite{liu2020few} & 65.86 & 60.57 & - & - & 80.61 & 67.35 & - & - \\
                                                    & TANE-G$^\star$ \cite{huang2022task} & 68.11 & 72.41 & - & - & 83.12 & 79.85 & - & - \\
                                                    & SnatcherF \cite{jeong2021few} &      67.23 &         70.10 &       69.74 &             58.02 &      82.00 &         76.57 &       76.97 &             61.64 \\
                                                    & OpenMax \cite{bendale2016towards} &      65.90 &         71.34 &       70.86 &             58.67 &      82.23 &         77.42 &       77.63 &             62.35 \\
                                                    & PROSER \cite{zhou2021learning} &      65.00 &         68.93 &       68.84 &             57.03 &      80.08 &         74.98 &       75.58 &             60.11 \\

                \toprule
                \multirow{5}{*}{Transductive classifiers} & LaplacianShot \cite{ziko2020laplacian} & 70.59 &         53.13 &       54.59 &             52.06 &  82.94 &         57.17 &       57.90 &             52.56 \\
                                                            & BDCSPN \cite{liu2020prototype} &      69.35 &         57.95 &       58.58 &             52.71 &    82.66 &         61.27 &       62.17 &             53.26 \\
                                                            & TIM-GD \cite{boudiaf2020transductive}  &      67.53 &         62.46 &       61.05 &             54.83 &      82.49 &         67.19 &       66.15 &             56.70 \\
                                                            & PT-MAP \cite{hu2021leveraging}  &      66.32 &         59.05 &       58.67 &             53.74 &      78.12 &         62.78 &       62.48 &             54.67 \\
                                                            & LR-ICI \cite{wang2020instance}  &      68.24 &         49.96 &       51.61 &             50.45 &      81.77 &         51.82 &       53.49 &             50.80 \\

                \midrule
                \rowcolor{lightgreen!25} Transductive Open-Set & \OSLO (ours) &  \textbf{71.73} &        \textbf{ 74.92} &     \textbf{ 74.61} &       \textbf{60.95}  &     \textbf{ 83.40} &   \textbf{82.59} &      \textbf{ 82.34} &       \textbf{66.98}\\
                \bottomrule
                \vspace{3pt} \\
                &  & \multicolumn{8}{c}{\textbf{\textit{tiered}-ImageNet}} \vspace{0.2em}\\
                
                & & \std{0.74} & \std{0.76} & \std{0.71} & \std{0.52} & \std{0.52} & \std{0.68} & \std{0.75} & \std{0.57} \vspace{0.2em}\\
                \toprule
                \multirow{6}{*}{OOD detection} & \knn \cite{knn_detector} &      -     &         73.97 &       73.15 &             60.74 &      -     &         80.22 &       80.06 &             65.47 \\
                                                & IForest \cite{iforest_detector}  &      -     &         54.57 &       54.24 &             51.85 &      -     &         62.31 &       60.82 &             54.72 \\
                                                & OCVSM \cite{ocsvm_detector}  &      -     &         71.22 &       71.17 &             58.81 &      -     &         71.20 &       68.23 &             61.09 \\
                                                & PCA \cite{pca_detector}  &      -     &         68.30 &       67.02 &             57.66 &      -     &         76.26 &       76.41 &             61.81 \\
                                                & COPOD \cite{copod_detector}  &      -     &         50.87 &       51.95 &             51.07 &      -     &         52.62 &       53.48 &             51.44 \\
                                                & HBOS                 &      -     &         57.54 &       56.67 &             52.98 &      -     &         60.91 &       59.95 &             54.15 \\

                \toprule
                \multirow{3}{*}{Inductive classifiers} & SimpleShot \cite{wang2019simpleshot} &      70.27 &         69.78 &       67.89 &             58.54 &      84.94 &         77.38 &       76.28 &             63.21 \\
                                                        & Baseline ++ \cite{Chen19} &      70.21 &         69.73 &       67.80 &             58.50 &      85.10 &         73.77 &       72.39 &             61.05 \\
                                                        & FEAT \cite{ye2020few} &      69.94 &         52.49 &       56.74 &             50.00 &      83.96 &         53.30 &       59.81 &             50.00 \\

                \midrule
                \multirow{4}{*}{Inductive Open-Set} & PEELER$^\star$ \cite{liu2020few} & 69.51 & 65.20 & - & - & 84.10 & 73.27 & - & - \\
                                                    & TANE-G$^\star$ \cite{huang2022task} & 70.58 & 73.53 & - & - & 85.38 & 81.54 & - & - \\
                                                    & SnatcherF \cite{jeong2021few}  &      69.94 &         74.02 &       73.33 &             60.79 &      83.96 &         81.90 &       81.67 &             66.89 \\
                                                    & OpenMax \cite{bendale2016towards} &      70.27 &         72.40 &       71.91 &             59.91 &      85.79 &         77.91 &       78.42 &             63.07 \\
                                                    & PROSER \cite{zhou2021learning} &      68.48 &         70.07 &       69.87 &             57.99 &      83.34 &         75.84 &       76.56 &             61.12 \\

                \toprule
                \multirow{5}{*}{Transductive classifiers} &  LaplacianShot \cite{ziko2020laplacian} &     75.66 &         57.82 &       58.41 &             53.67 &     86.23 &         63.75 &       63.65 &             55.36 \\
                                                            & BDCSPN \cite{liu2020prototype} &      74.07 &         62.13 &       61.84 &             54.53 &      85.65 &         67.41 &       67.57 &             56.30 \\
                                                            & TIM-GD \cite{boudiaf2020transductive}  &      72.56 &         68.08 &       65.97 &             57.84 &      85.70 &         74.67 &       73.06 &             61.59 \\
                                                            & PT-MAP \cite{hu2021leveraging}  &      71.13 &         64.48 &       62.94 &             56.25 &      82.81 &         71.08 &       69.89 &             59.11 \\
                                                            & LR-ICI \cite{wang2020instance} &      73.80 &         49.32 &       51.41 &             50.35 &      85.21 &         51.65 &       53.85 &             50.79 \\

                \midrule
                \rowcolor{lightgreen!25} Transductive Open-Set & \OSLO (ours) &     \textbf{76.64} &         \textbf{79.06} &      \textbf{ 79.07} &             \textbf{64.36} &     \textbf{86.35} &  \textbf{86.92} &       \textbf{87.28} &            \textbf{ 71.98} \\
                \bottomrule
            \end{tabular}
        }
        \label{tab:benchmark_results}
    \end{table*}

        \begin{figure*}[t]
            \centering
            \includegraphics[width=\linewidth]{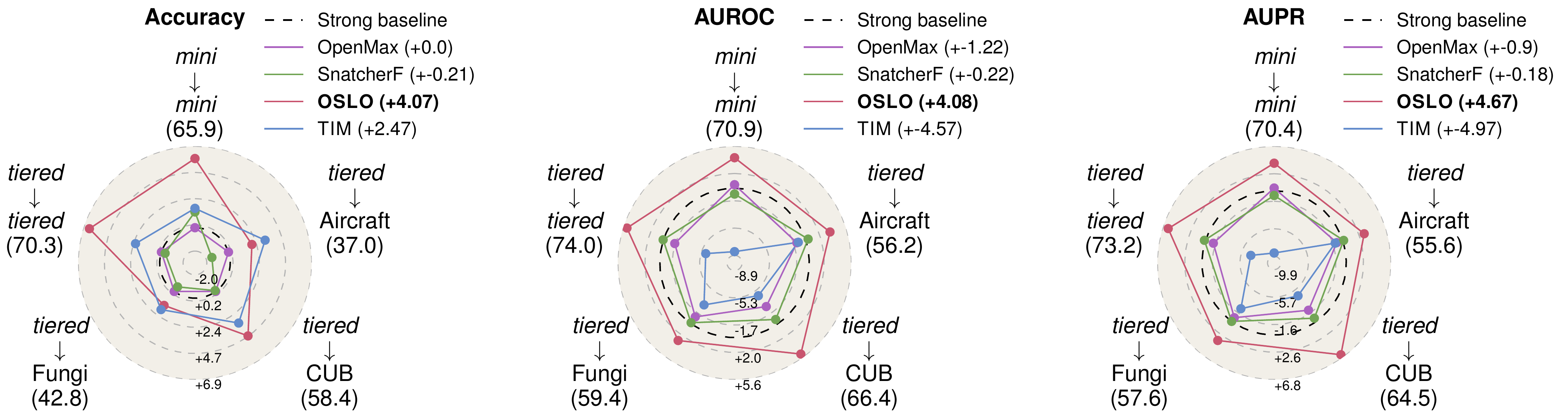} \\
            \caption{\textbf{\OSLO improves open-set performances on a wide variety of tasks.} Relative 1-shot performance of the best methods of each family w.r.t the \textit{Strong baseline} using a ResNet-12, across a set of 5 scenarios, including 3 with domain-shift. Each vertex represents one scenario, e.g. \textit{tiered}$\rightarrow$Fungi ($x$) means the feature extractor was pre-trained on \textit{tiered}-ImageNet, test tasks are sampled from Fungi, and the \textit{Strong Baseline} performance is $x$. For each method, the average relative improvement across the 5 scenarios is reported in parenthesis in the legend. The same charts are provided in the supplementary materials for the 5-shot setting and using a WideResNet backbone.}
            \label{fig:spider_charts}
        \end{figure*}

        \magicpar{Baselines.} One goal of this work is to fairly evaluate different strategies to address the FSOSR problem. In particular, we benchmark 4 families of methods: (i) popular Outlier Detection methods, e.g. Nearest-Neighbor \cite{knn_detector}, (ii) Inductive Few-Shot classifiers, e.g. SimpleShot \cite{wang2019simpleshot} (iii) Inductive Open-Set methods formed by standard methods such as OpenMax \cite{bendale2016towards} and Few-Shot methods such as Snatcher \cite{jeong2021few} (iv) Transductive classifiers, e.g. TIM \cite{boudiaf2020transductive}, that implicitly rely on the closed-set assumption, and finally (v) Transductive Open-Set introduced in this work through \OSLO. Following \cite{jeong2021few}, closed-set few-shot classifiers are turned into open-set classifiers by considering the negative of the maximum probability as a measure of outlierness. Furthermore, we found that applying a center-normalize transformation $\psi_{\Mubf}: \xbf \mapsto (\xbf - \Mubf) / ||\xbf - \Mubf ||_2 $ on the features extracted by $\phi_{\thetabf}$ benefited all methods. Therefore, we apply it to the features before applying any method, using an inductive \textit{Base centering} \cite{wang2019simpleshot} for inductive methods $\Mubf_{Base}=\frac{1}{|\mathcal{D}_{base}|} \sum_{\xbf \in \mathcal{D}_{base}} \phi_{\thetabf}(x)$, and a transductive \textit{Task centering} \cite{hu2021leveraging} ${\Mubf_{Task}=\frac{1}{|\support \cup \query|} \sum_{\xbf \in \support \cup \query} \phi_{\thetabf}(x)}$ for all transductive methods. Since features are normalized, we empirically found it beneficial to re-normalize centroids $\mubf_k \leftarrow \mubf_k / ||\mubf_k||_2$ after each update from Prop. \ref{prop:oslo}, which we show in the Appendix remains a valid minimizer of \cref{eq:oslo_objective} when adding the constraint $||\mubf_k||_2=1$.
    
        \magicpar{Hyperparameters.} For all methods, we define a grid over salient hyper-parameters and tune over the validation split of \textit{mini}-ImageNet. To avoid cumbersome per-dataset tuning, and evaluate the generalizability of methods, we then keep hyper-parameters fixed across all other experiments. 
        
        \magicpar{Architectures and checkpoints.} To provide the fairest comparison, all non-episodic methods are tuned and tested using off-the-shelf pre-trained checkpoints. All results except \autoref{fig:barplots} are produced using the pre-trained ResNet-12 and Wide-ResNet 28-10 checkpoints provided by the authors from \cite{ye2020few}. As for episodically-finetuned models required by Snatcher \cite{jeong2021few} and FEAT \cite{ye2020few}, checkpoints are obtained from the authors' respective repositories. Finally, to challenge the model-agnosticity of our method, we resort to an additional set of 10 ImageNet pre-trained models covering three distinct architectures: ResNet-50 \cite{resnet} for CNNs, ViT-B/16 \cite{vit} for vision transformers, and Mixer-B/16 \cite{mlp_mixer} for MLP-Mixer. These models are taken from the excellent \textsc{TIMM} library \cite{rw2019timm}.

        \magicpar{Datasets and tasks.} We experiment with a total of 5 vision datasets. As standard FSC benchmarks, we use the \textit{mini}-ImageNet \cite{Vinyals16} dataset with 100 classes and the larger \textit{tiered}-ImageNet \cite{tiered_imagenet} dataset with 608 classes. We also experiment on more challenging cross-domain tasks formed by using 3 finer-grained datasets: the Caltech-UCSD Birds 200 \cite{cub} (CUB) dataset, with 200 classes, the FGVC-Aircraft dataset \cite{maji2013fine} with 100 classes, and the Fungi classification challenge \cite{schroeder2018fgvcx} with 1394 classes. Following standard FSOSR protocol, support sets contain $|\cseen|=5$ closed-set classes with 1 or 5 instances, or \textit{shots}, per class, and query sets are formed by sampling 15 instances per class, from a total of ten classes: the five closed-set classes and an additional set of $|\cunseen|=5$ open-set classes. We follow this setting for a fair comparison with previous works \cite{jeong2021few, liu2020few} which sample open-set query instances from only 5 classes. We also report results in supplementary materials for a more general setting in which open-set query instances are sampled indifferently from all remaining classes in the test set.
        
    \subsection{Results}

        \magicpar{Simplest inductive methods are competitive.} 
        The first surprising result comes from analyzing the performances of standard OOD detectors on the FSOSR problem. Fig. \ref{tab:benchmark_results} shows that \knn~ and \textsc{PCA} outperform, by far, arguably more advanced methods that are OCVSM and Isolation Forest. This result contrasts with standard high-dimensional benchmarks \cite{zhao2019pyod} where \knn~ falls typically short of the latter, indicating that the very difficult challenge posed by FSOSR may lead advanced methods to overfit. In fact, Fig. \ref{fig:spider_charts} shows that across 5 scenarios, the combination SimpleShot \cite{wang2019simpleshot}+ \knn~ \cite{knn_detector} formed by the simplest FS-inductive classifier and the simplest inductive OOD detector is a strong baseline that outperforms all specialized open-set methods. We refer to this combination as \textit{Strong baseline} in Figures \ref{fig:spider_charts} and \ref{fig:barplots}. Additional results for the Wide-ResNet architecture are provided in the supplementary material.
        
    \begin{figure*}[t]
        \centering
        \includegraphics[width=0.99\linewidth]{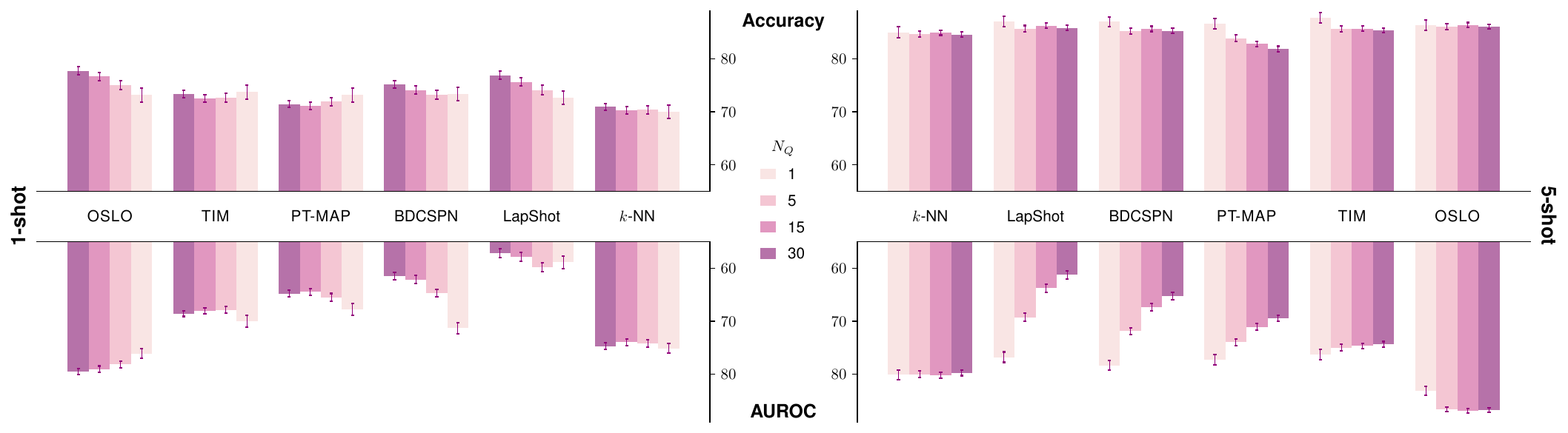}
        \caption{\textbf{\OSLO improves performance even with few queries.} We study the closed-set (accuracy) and open-set (AUROC) performance of transductive methods depending on the size of the query set on \textit{tiered}-ImageNet in the 1-shot and 5-shot settings. The total size $|Q|$ of the query set is obtained by multiplying the number of queries per class $N_Q$ by the number of classes in the task (\ie 5) and adding as many outlier queries \eg $N_Q = 1$ corresponds to 1 query per class and 5 open-set queries \ie $|Q| = 10$. We add the inductive method $k$-NN + SimpleShot to compare with a method that is by nature independent of the number of queries. The results for \textit{mini}-ImageNet are provided in the supplementary materials.}
        \label{fig:variate-query-tiered}
    \end{figure*}

        \magicpar{Transductive methods still improve accuracy but degrade outlier detection.} As shown in Table \ref{tab:benchmark_results}, most transductive classifiers still offer a significant boost in closed-set accuracy, even in the presence of outliers in the query set. Note that this contrasts with findings from the semi-supervised literature, where standard methods drop below the baseline in the presence of even a small fraction of outliers \cite{yu2020multi,chen2020semi,saito2021openmatch, killamsetty2021retrieve}. We hypothesize that the deliberate under-parametrization of few-shot methods --typically only training a linear classifier--, required to avoid overfitting the support set, partly explains such robustness. However, transductive methods still largely underperform in outlier detection, with AUROCs as low as 52 \% (50\% being a random detector) for LaplacianShot. Note that the \textit{outlierness} score for these methods is based on the negative of the maximum probability, therefore this result can be interpreted as transductive methods having artificially matched the prediction confidence for outliers with the confidence for inliers.

        \magicpar{\OSLO achieves the best trade-off.} Benchmark results in Fig. \ref{tab:benchmark_results} show that \OSLO surpasses the best transductive methods in terms of closed-set accuracy, while consistently outperforming existing out-of-distribution and open-set detection competitors on outlier detection ability. Interestingly, while the gap between closed-set accuracy of transductive methods and inductive ones typically contracts with more shots, the outlier detection performance of \OSLO remains largely superior to its inductive competitors even in the 5-shot scenario, where a consistent 3-6\% gap in AUROC and AUPR with the second-best method can be observed. We accumulate further evidence of \OSLO's superiority by introducing 3 additional cross-domain scenarios in Fig. \ref{fig:spider_charts}, corresponding to a base model pre-trained on \textit{tiered}-ImageNet, but tested on CUB, Aircraft, and Fungi datasets. In such challenging scenarios, where both feature and class distributions shift, \OSLO remains competitive in closed-set accuracy and largely outperforms other methods in outlier detection.

        \magicpar{\OSLO benefits from more query samples.} A critical question for transductive methods is the dependency of their performance on the size of the query set. Intuitively, a larger query set will provide more unlabeled data and thus lead to better results. We exhibit this relation in \autoref{fig:variate-query-tiered} by spanning the number of queries per class from 1 to 30. We observe that the closed-set accuracy of most transductive methods is stable across this span in the 5-shot scenario. In the 1-shot scenario, \OSLO gains from additional queries but stays above the baseline even with a small number of queries. Interestingly enough, \OSLO is the only transductive method to improve its outlier detection ability when the number of queries increases. 

          \begin{figure*}[t]
            \centering
            \includegraphics[width=0.75\linewidth]{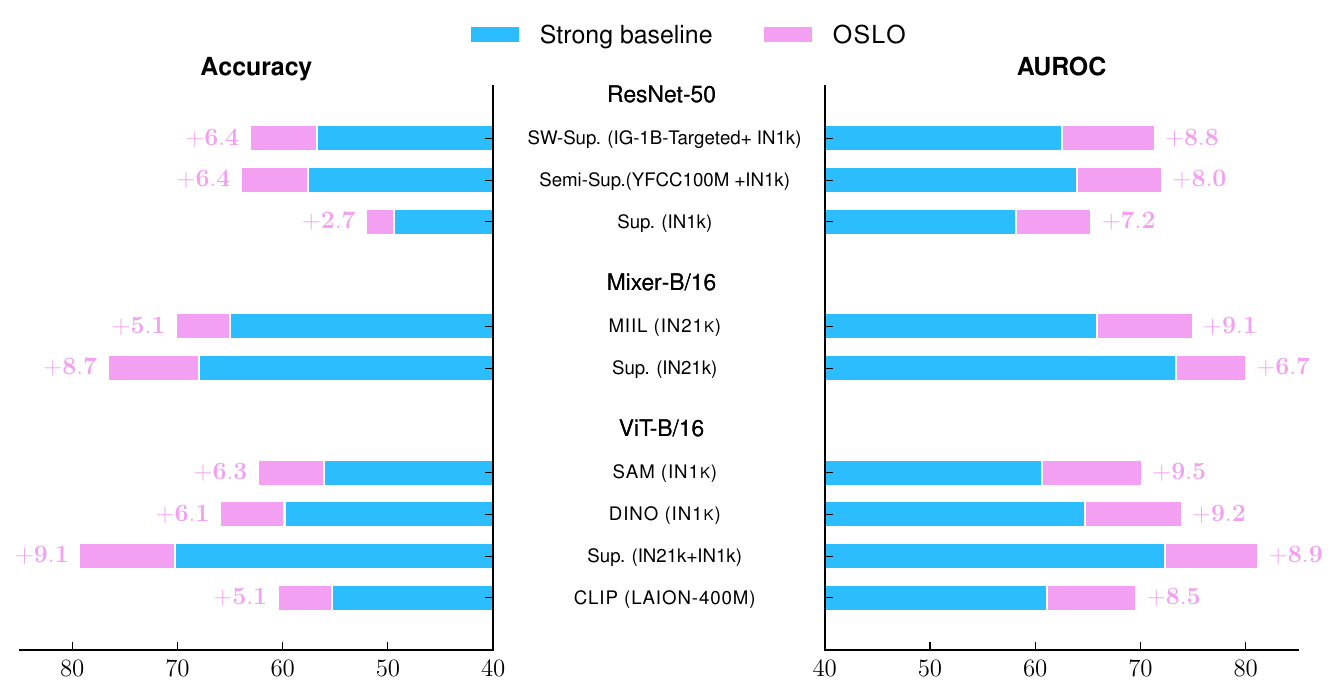}
            \caption{\textbf{\OSLO's improvement is consistent across many architectures and training strategies.} To evaluate model-agnosticity, we compare \OSLO to the Strong baseline on challenging 1-shot Fungi tasks. We experiment across 3 largely distinct architectures: ResNet-50 (CNN) \cite{resnet}, ViT-B/16 (Vision Transformer) \cite{vit}, and Mixer-B/16 (MLP-Mixer) \cite{mlp_mixer}. For each architecture, we include different types of pre-training, including Supervised (Sup.), Semi-Supervised, Semi-Weakly Supervised (SW Sup.) \cite{semi_sup}, \textsc{DINO} \cite{dino}, \textsc{SAM} \cite{sam}, \textsc{MIIL} \cite{miil}. Improvements over the baseline are consistently significant and generally higher than those observed with the ResNet-12 in \autoref{fig:spider_charts}.}
            \label{fig:barplots}
        \end{figure*}  
        \begin{table}[t]
            \centering
            \small
            \caption{\textbf{\OSLO's ablation study} along two factors described in \autoref{sec:ablation_study}. Results are produced on the 1-shot scenario on \textit{mini}-ImageNet, with a ResNet-12. \\}
            \vspace{-12pt}
            \resizebox{.35\textwidth}{!}{
                \begin{tabular}{lcc}
                    \toprule
                    (i) \textit{Inlierness} latent & Acc & AUROC \\
                    \midrule
                    Ignore \eqref{eq:standard_likelihood} & 69.42 & 64.97 \\
                    Leverage \eqref{eq:open_likelihood} & 71.73 & 74.92 \\
                    \toprule
                    (ii) Optimization steps & Acc & AUROC \\
                    \midrule
                    At initialization & 66.63 & 71.76 \\
                    After optimization & 71.73 & 74.92 \\
                    \bottomrule
                \end{tabular}
            }
            \label{tab:ablation_study}
        \end{table} 
        
        \magicpar{\OSLO steps toward model-agnosticity.} 
        We evaluate \OSLO's \textit{model-agnosticity} by its ability to maintain consistent improvement over the \textit{Strong Baseline}, regardless of the model used, and without hyperparameter adjustment.  In that regard, we depart from the standard ResNet-12 and cover 3 largely distinct architectures, each encoding different inductive biases. To further strengthen our empirical demonstration of \OSLO's model-agnosticity, for each architecture, we consider several training strategies spanning different paradigms -- unsupervised, supervised, semi and semi-weakly supervised -- and using different types of data --image, text--. Results in \autoref{fig:barplots} show the relative improvement of \OSLO w.r.t the strong baseline in the 1-shot scenario on the $\ast \rightarrow$ Fungi benchmark. Without any tuning, \OSLO remains able to leverage the strong expressive power of large-scale models, and even consistently widens the gap with the strong baseline, achieving a remarkable performance of 79\% accuracy and 81\% AUROC with the ViT-B/16. This set of results testifies to how easy obtaining highly competitive results on difficult specialized tasks can be by combining \OSLO with the latest models. 


    \magicpar{Ablation study.} \label{sec:ablation_study}
    We perform an ablation study on the important ingredients of \OSLO. As a core contribution of our work, we show in \cref{tab:ablation_study} that the presence \emph{and} optimization of the latent \textit{inlierness} scores is crucial. In particular, the closed-form latent score $\xi$ yields strong outlier recognition performance, even at \textit{initialization} (i.e. after the very first update from Prop. \ref{prop:oslo}). Interestingly, refining the parametric model without accounting for $\xibf$ in $\mathbf{Z}$ and $\mubf$'s updates (i.e. standard likelihood) allows the model to fit those outliers, leading to significantly worse outlier detection, from $71.76 \%$ to $64.97\%$. On the other hand, accounting for $\xibf$, as proposed in \OSLO, improves the outlier detection by more than $3\%$ over the initial state, and closed-set accuracy by more than $5 \%$. In the end, in a fully apples-to-apples comparison, \OSLO outperforms its standard likelihood counterpart by more than $2 \%$ in accuracy and $10\%$ in outlier detection. We strengthen this ablation study in the Appendix.

\section{Discussion}
\label{sec:discussions}

    \magicpar{Limitations.} Unlike inductive methods,
    transductive methods are inevitably affected by the amount of
    unlabelled data provided, which in real-world scenarios cannot
    necessarily be controlled. \OSLO is no exception, and fewer query
    samples tend to decrease its performance. In the extreme case with
     only 1 sample per class, \OSLO's performance comes close to our
     inductive baseline. In those scenarios where unlabelled data is
     particularly scarce, the benefits brought by transduction remain
     therefore limited. As a second limitation, poorer representations
     appear to diminish \OSLO's competitive advantage in closed-set
     accuracy. In particular, \OSLO's closed-set accuracy stands more
     than $6\%$ above the baseline on the $\textit{tiered} \rightarrow \textit{tiered}$
     scenario but reduces to ~$2\%$ in the most challenging-domain scenario
     $\textit{tiered} \rightarrow \textit{Aircraft}$. \cref{fig:barplots} further corroborates this hypothesis, with \OSLO 's accuracy outperforming the baseline's by $9\%$ with the best transformer, but by only $2.7\%$ on the least performing model.

    \magicpar{Conclusion.} We presented \OSLO, the first transductive method for FSOSR. \OSLO extends the vanilla maximum likelihood objective in two important ways, First, it accounts for the constraints imposed by the provided supervision. More importantly, it explicitly models the potential presence of outliers in its very latent model, allowing it to co-learn the optimal closed-set model and outlier assignments. Beyond FSOSR, we believe \OSLO presents a general, conceptually simple, and completely modular formulation to leverage unlabelled data in the potential presence of outliers. That, of course, naturally extends to other classification settings, such as large-scale open-set detection, but to other tasks as well, such as segmentation in which \textit{background} pixels could be viewed as \textit{outliers} with respect to closed-set classes. We hope \OSLO inspires further work in that direction.


{\small
\bibliographystyle{ieee_fullname}
\bibliography{egbib}
}

\clearpage
\appendix

\section{Proof of Proposition \ref{prop:oslo}}

 \magicpar{Reminder of Proposition \ref{prop:oslo}}
 OSLO's optimization problem, being defined in \eqref{eq:oslo_objective} as: 
     \begin{align*}
            \max_{\mubf, \Zbf, \xibf} & \quad \mathcal{L}_{\text{O}}(\Zbf, \xibf, \mubf) + \mathcal{L}_{\text{soft}}(\Zbf, \xibf) \nonumber \\ 
            \text{s.t } & \quad \zbf_i \in \Delta^K, \quad \xi_i \in [0, 1] \quad \forall ~ i\\
            & \quad \zbf_i = \ybf_i, \quad \xi_i = 1,  \quad i \leq |\support|   \nonumber
    \end{align*}
 can be minimized by alternating the following updates:
        \begin{align*}
            \xi_i ^{(t+1)} &=
                \begin{cases}
                    1 \quad \text{if } i \leq |\support| \\
                    \sigma \left(\displaystyle\frac{1}{\lambda_\xi} \displaystyle \sum_{k=1}^K z_{ik}^{(t)} \log p(\xbf_i|k; \mubf^{(t)}))\right) \text{else}
                \end{cases} \\
            \zbf_i^{(t+1)} &\propto 
                \begin{cases}
                    \ybf_i \quad \text{if } i \leq |\support| \\
                    \exp \left(\displaystyle\frac{\xi_i^{(t+1)}}{\lambda_z} \log p(\xbf_i|~\cdot~; \mubf^{(t)}) \right)  \text{else}
                \end{cases} \\
            \mubf^{(t+1)}_k &= \frac{1}{\displaystyle\sum_{i=1}^{|\support|+ |\query|} \xi_i^{(t+1)} z_{ik}^{(t+1)}} \sum_{i=1}^{|\support| + |\query|} \xi_i^{(t+1)} z_{ik}^{(t+1)}\phi_{\theta}(\xbf_i)
        \end{align*}
        where $\sigma$ denotes the sigmoid operation. 

\begin{proof}
    We denote by $\nabla_{\cdot} (\mathcal{L}_{\text{O}} + \mathcal{L}_{\text{soft}})$ the partial derivative of \OSLO's optimization problem. We calculate the updates of $\xi_i$ and $z_{ik}$ for $i > |S|$, and of $\mubf_k$, by finding the annulation point of their partial derivative.

    \begin{align*}
        &\nabla_{\xi_i} (\mathcal{L}_{\text{O}} + \mathcal{L}_{\text{soft}}) = 0 \\
            \Leftrightarrow \quad &
                \sum_{k=1}^K z_{ik} \log \left(p(\xbf_i, k; \mubf) \right) \\
                & \quad =  \lambda_\xi \left((\log\xi_i + 1) - (\log(1-\xi_i) +1)\right) \\
            \Leftrightarrow \quad &
                \frac{1}{\lambda_\xi} \sum_{k=1}^K z_{ik} \log \left(p(\xbf_i, k; \mubf) \right)
                = \log \left( \frac{\xi_i}{1- \xi_i} \right) \\
            \Leftrightarrow \quad  & \xi_i = \sigma \left( \frac{1}{\lambda_\xi}  \sum_{k=1}^K z_{ik} \log \left(p(\xbf_i, k; \mubf) \right) \right) 
    \end{align*}
    
    \begin{align*}
         &\nabla_{z_{ik}} (\mathcal{L}_{\text{O}} + \mathcal{L}_{\text{soft}}) = 0 \\
            \Leftrightarrow \quad & \xi_i \log \left(p(\xbf_i, k; \mubf) \right) = \lambda_z \left( \log z_{ik} + 1 \right) \\
            \Rightarrow \quad & z_{ik} \propto \exp \left( \frac{\xi_i}{\lambda_z} \log \left(p(\xbf_i, k; \mubf) \right) \right)
    \end{align*}

    \begin{align*}
         &\nabla_{\mubf_k} (\mathcal{L}_{\text{O}} + \mathcal{L}_{\text{soft}}) = 0 \\
            \Leftrightarrow \quad & \sum_{i=1}^{|\support| + |\query|} \xi_i z_{ik} \left( \phi_{\thetabf} (\xbf_i) - \mubf_k \right) = 0 \\
            \Leftrightarrow \quad & \mubf_k =  \dfrac{\sum_{i=1}^{|\support| + |\query|} \xi_i z_{ik} \phi_{\thetabf} (\xbf_i)}{\sum_{i=1}^{|\support| + |\query|} \xi_i z_{ik}}
    \end{align*}
\end{proof}

\section{Normalizing centroids}

    Because we work with normalized features, we state in our implementation details that we found normalizing $||\mubf||$ after each update helps. Here we show that this "projected step" is actually the exact solution to the optimization problem \cref{eq:oslo_objective} when adding the constraint $\mubf \in \unitball$, where $\unitball = \{\xbf: ||\xbf||_2=1 \}$ is the unit hypersphere. 
    
    Specifically, adding the constraint $\mubf \in \unitball$ modifies the Lagrangian by infinitely penalizing $\mubf_k$ for being outside the unit hypersphere. Without loss of generality, we only consider the part of the Lagrangian pertaining to $\mubf_k$ for some $k \in [1, K]$, which we refer to as $\mathcal{L}_k$:

    \begin{align*}
        \mathcal{L}_k(\mubf_k) = \sum_{i=1}^{|\support|+|\query|} \xi_i z_{ik} ||\mubf_k - \phi_\theta(\xbf_i)||^2 + \mathcal{L}_{\unitball}(\mubf_k)
    \end{align*}
    where $\mathcal{L}_{\unitball}(\mubf_k)$ equals 0 if $\mubf_k \in \unitball$ and $\infty$ otherwise. Because $\mathcal{L}_k$ is no longer differentiable, we introduce the subdifferential operator $\partial_{\cdot}(\cdot)$, which generalizes the standard notion of differentiability. Akin to the standard case, we look for $\mubf_k$ such that:
    \begin{align*}
        0 \in \partial_{\mubf_k} \mathcal{L}_k(\mubf_k) ,
    \end{align*}
    which amounts to:
    \begin{align*}
        \Leftrightarrow& \quad 0 \in \{\nabla_{\mubf_k} \sum_{i=1}^{|\support|+|\query|} \xi_i z_{ik} ||\mubf_k - \phi_\theta(\xbf_i)||^2\} + \partial_{\mubf_k} \mathcal{L}_{\unitball}(\mubf_k) \\
        \Leftrightarrow& \quad \sum_{i=1}^{|\support|+|\query|} \xi_i z_{ik} ~ \phi_{\theta}(\xbf_i) - \mubf_k (\sum_{i=1}^{|\support|+|\query|} \xi_i z_{ik}) \in \partial_{\mubf_k} \mathcal{L}_{\unitball}(\mubf_k) \\
        \Leftrightarrow& \quad \frac{\sum_{i=1}^{|\support|+|\query|} \xi_i z_{ik} ~ \phi_{\theta}(\xbf_i)}{\sum_{i=1}^{|\support|+|\query|} \xi_i z_{ik}} - \mubf_k \in \partial_{\mubf_k} \frac{1}{\sum_{i=1}^{|\support|+|\query|} \xi_i z_{ik}}\mathcal{L}_{\unitball}(\mubf_k) \\
        \Leftrightarrow& \quad \frac{\sum_{i=1}^{|\support|+|\query|} \xi_i z_{ik} ~ \phi_{\theta}(\xbf_i)}{\sum_{i=1}^{|\support|+|\query|} \xi_i z_{ik}} - \mubf_k \in \partial_{\mubf_k} \mathcal{L}_{\unitball}(\mubf_k) \\
        \Leftrightarrow& \quad \mubf_k = \text{Proj}_{\unitball}(\frac{\sum_{i=1}^{|\support|+|\query|} \xi_i z_{ik} ~ \phi_{\theta}(\xbf_i)}{\sum_{i=1}^{|\support|+|\query|} \xi_i z_{ik}}) \\
    \end{align*}
    where the penultimate step holds because $\lambda \mathcal{L}_{\unitball}(\mubf_k) = \mathcal{L}_{\unitball}(\mubf_k)$ by definition of $\mathcal{L}_{\unitball}(\mubf_k)$, and the last step holds because the projection operator $\text{Proj}_{\unitball}(\mubf_k) = \frac{\mubf_k}{||\mubf_k||}$ is the proximity operator of the constraint function $\mathcal{L}_{\unitball}(\mubf_k)$.

\section{Metrics}

Here we provide some details about the metrics used in Section \ref{sec:experiments}

    \textbf{Acc}: the classification accuracy on the closed-set instances of the query set (\ie $y^q \in \mathbb{C_S}$).
    
    \textbf{AUROC}: the area under the ROC curve is an almost mandatory metric for any OOD detection task. For a set of outlier predictions in $[O,1]$ and their ground truth ($0$ for inliers, $1$ for outliers), any threshold $\gamma \in [O,1]$ gives a true positive rate $\textit{TP}(\gamma)$ (\ie recall) and a false positive rate $\textit{FP}(\gamma)$. By rolling this threshold, we obtain a plot of \textit{TP} as a function of \textit{FP} \ie the ROC curve. The area under this curve is a measure of the discrimination ability of the outlier detector. Random predictions lead to an AUROC of $50\%$.
    
    \textbf{AUPR}: the area under the precision-recall (PR) curve is also a common metric in OOD detection. With the same principle as the ROC curve, the PR curve plots the precision as a function of the recall. Random predictions lead to an AUPR equal to the proportion of outliers in the query set \ie $50\%$ in our set-up.
    
    \textbf{Prec@0.9}: the precision at $90\%$ recall is the achievable precision on the few-shot open-set recognition task when setting the threshold allowing a recall of $90\%$ for the same task. While AUROC and AUPR are global metrics, \textit{Prec@0.9} measures the ability of the detector to solve a specific problem, which is the detection of almost all outliers (\eg for raising an alert when open-set instances appear so a human operator can create appropriate new classes). Since all detectors are able to achieve high recall with a sufficiently permissive threshold $\gamma$, an excellent way to compare them is to measure the precision of the predictor at a given level of recall (\ie the proportion of false alarms that the human operator will have to handle). Random predictions lead to a \textit{Prec@0.9} equal to the proportion of outliers in the query set \ie $50\%$ in our set-up.

\section{Effects of the inlier latent on closed-set model parameters}

\begin{figure*}
    \centering
    \begin{subfigure}[b]{\textwidth}
        \centering
        \includegraphics[width=.7\textwidth]{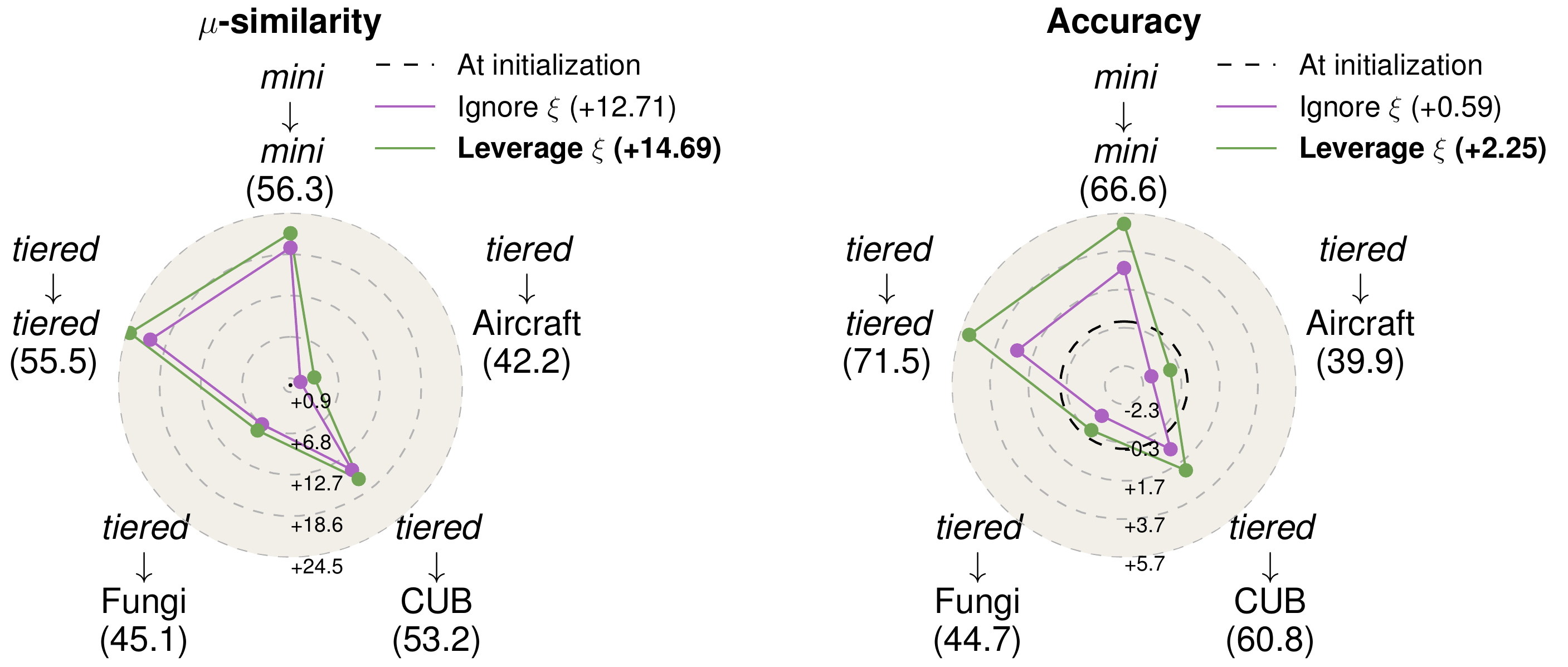}
        \caption{1-shot}
    \end{subfigure}
    \begin{subfigure}[b]{\textwidth}
        \centering
        \includegraphics[width=.7\textwidth]{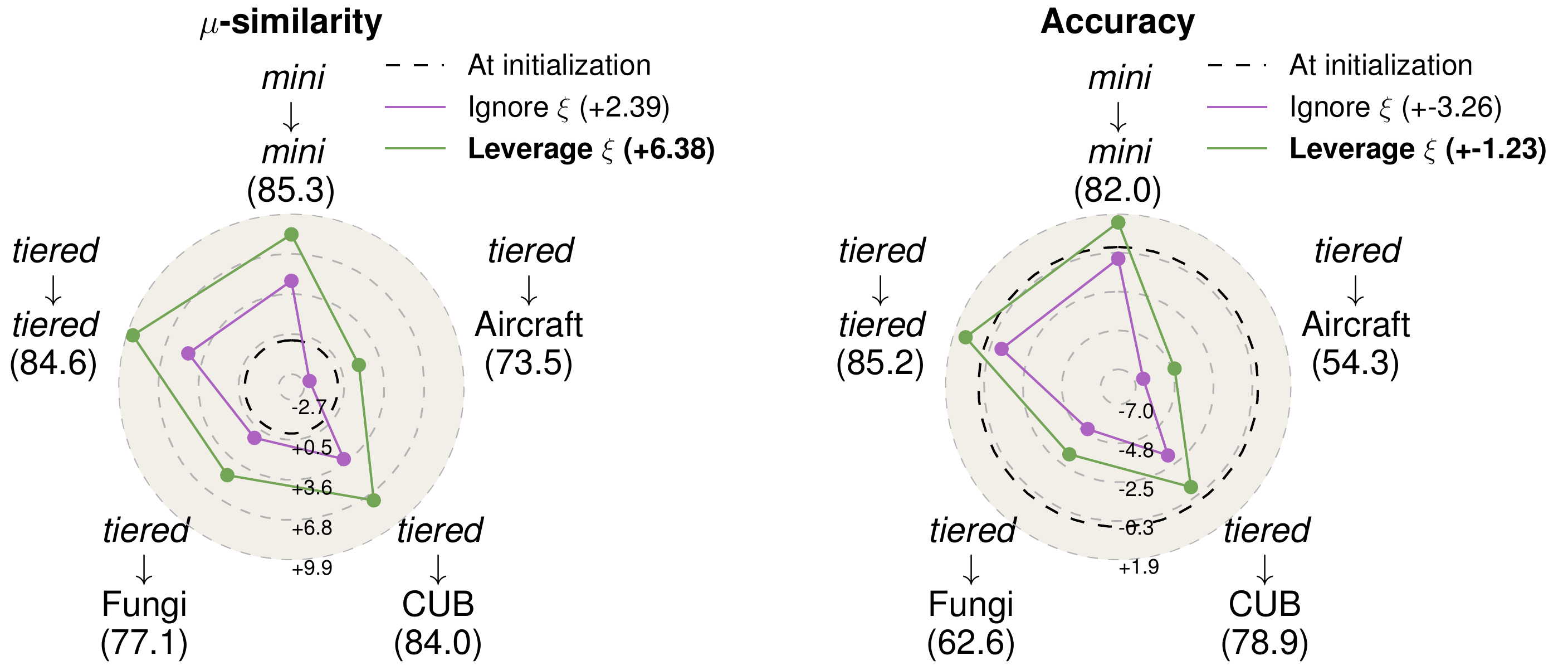}
        \caption{1-shot}
    \end{subfigure}
    \caption{Effects of leveraging the inlier latent $\xibf$ on the quality of the closed-set parameters $\Zbf$ (measured with the accuracy) and $\mubf$ (measured with the cosine similarity between $\mubf$ and the ground truth prototypes computed as the average of all support and query embeddings for each class). We compare the full \OSLO method from Eq. \eqref{eq:open_likelihood} (Leverage $\xibf$) with the standard likelihood objective from Eq. \eqref{eq:standard_likelihood} (Ignore $\xibf$) and no optimization (At initialization). This figure follow the same logic as Figure. \ref{fig:spider_charts}.}
    
    \label{fig:latent-effects}
\end{figure*}

We reported in Tab. \ref{tab:ablation_study} an ablation on the effect of introducing $\xibf$ (Eq. \eqref{eq:open_likelihood}) on the obtained $\Zbf$ (latent class assignments). Here we go further into this ablation by illustrating in Figure \ref{fig:latent-effects} how leveraging $\xibf$ yields better estimates of both $\Zbf$ the prototypes $\mubf$. The latter is measured by the similarity between $\mubf$ obtained after optimization and the ground-truth prototypes (using the support and query labels of each task). These results indicate that leveraging the inlier latent consistently improves the parametric model $\mubf$ across all benchmarks. Interestingly, this does not result in better latent class-assignments $\Zbf$ in the cross-domain scenarios.

\section{Broad Open-Set setting}

As we state in Section \ref{sec:experiments}, in the standard FSOSR setting \cite{jeong2021few, liu2020few}:
\begin{itemize}
    \item support sets contain $|\cseen|=5$ closed-set classes with 1 or 5 instances, or \textit{shots}, per class;
    \item query sets are formed by sampling 15 instances per class, from a total of ten classes: 
    \begin{itemize}
        \item the five closed-set classes $\cseen$;
        \item an additional set of $|\cunseen|=5$ open-set classes.
    \end{itemize}
\end{itemize}

This is a very strong assumption on the distribution of open-set samples. While this will not affect an inductive method, it is likely to impact the performance of transductive methods like \OSLO. In this section, we provide additional results in a more realistic setting. In this new setting, the query set is formed by sampling 15 instances for each of the 5 closed-set classes, plus $5 \times 15 = 75$ open-set instances, which are sampled indifferently from all remaining classes in the test set.

\begin{figure*}
        \centering
        \includegraphics[width=.99\textwidth]{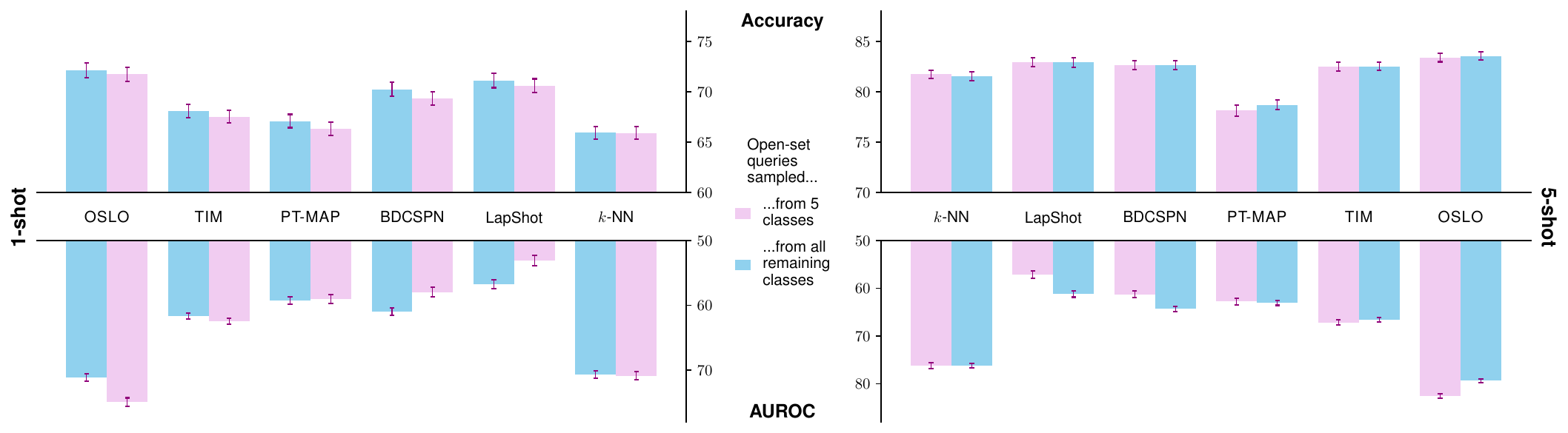}
        \caption{\textbf{Performance of transductive methods in the broad open-set setting.} We study the closed-set (accuracy) and open-set (AUROC) performance of transductive methods depending on the size of the query set on \textit{mini}-ImageNet in the 1-shot and 5-shot settings. We add the inductive method $k$-NN + SimpleShot to compare with a method that is by nature independent to the number of queries.}
        \label{fig:broad}
    \end{figure*}

Results in Figure \ref{fig:broad} show that the distribution of open-set queries is indeed a major factor in both closed-set and open-set performances for most transductive methods. Interestingly enough, some methods like Laplacian Shot \cite{ziko2020laplacian} or BDCSPN \cite{liu2020prototype} benefit from this relaxation of the previous open-set assumption. However, while \OSLO's closed-set accuracy increases in the new setting, its open-set recognition ability decreases (while still achieving the best results across the benchmark).

\section{The difficulty of FSOSR}

        As stated in Section \ref{sec:fsosr_setting}, our method follows the model-agnostic setting. Therefore, we perform Few-Shot Open-Set Recognition on features lying in a feature space $\featurespace$ and extracted by a frozen model $\phi_{\thetabf}: \inputspace \rightarrow \featurespace$, whose parameters $\thetabf$ were trained on some large dataset $\mathcal{D}_{base} = \{(\xbf^b_i,y^b_i)\}_{i=1...|\mathcal{D}_{base}|}$ such that for all $i$, $y^b_i \in \mathbb{C}_{base}$ with $\mathbb{C}_{base} \cap \cseen = \mathbb{C}_{base} \cap \cunseen = \emptyset$.

        While model-agnosticity is a very strong selling point for a few-shot learning method, it also comes with very difficult challenges, especially for an outlier detection task. In this section, we aim at providing a better understanding of the difficulty of FSOSR with both a qualitative and quantitative study of the clusters formed by novel classes $\cunseen$ when embedded by a feature extractor $\phi_{\thetabf}$ untrained on $\cunseen$.

        \begin{figure*}
            \centering
            \includegraphics[width=.44\textwidth]{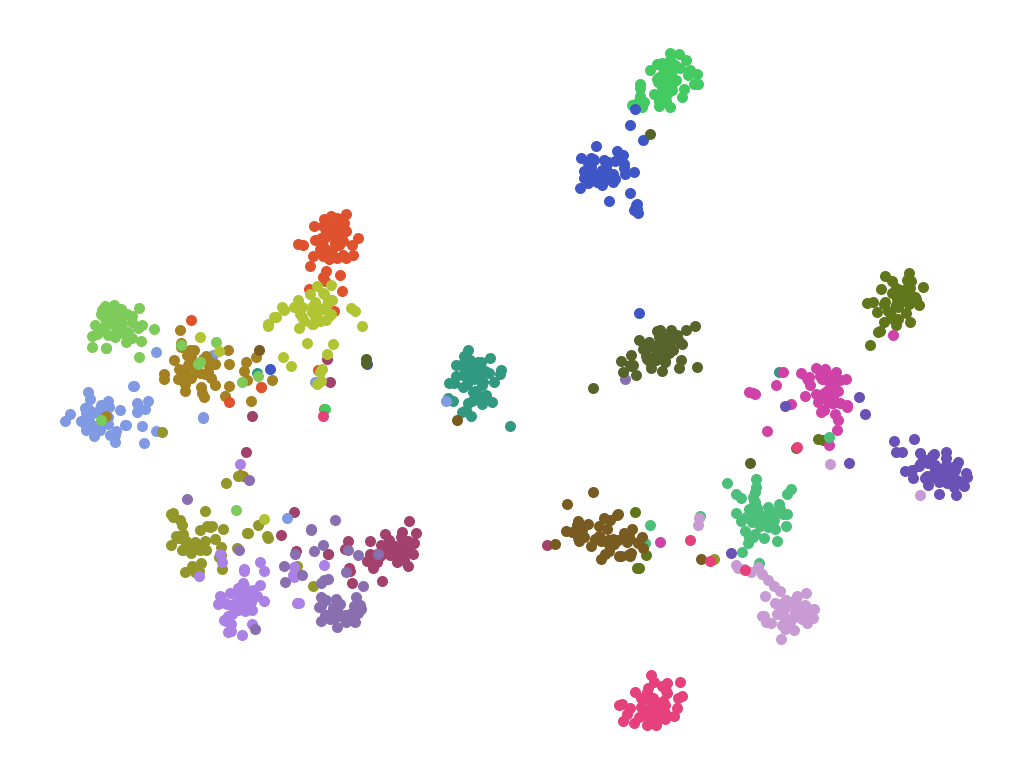}
            \includegraphics[width=.44\textwidth]{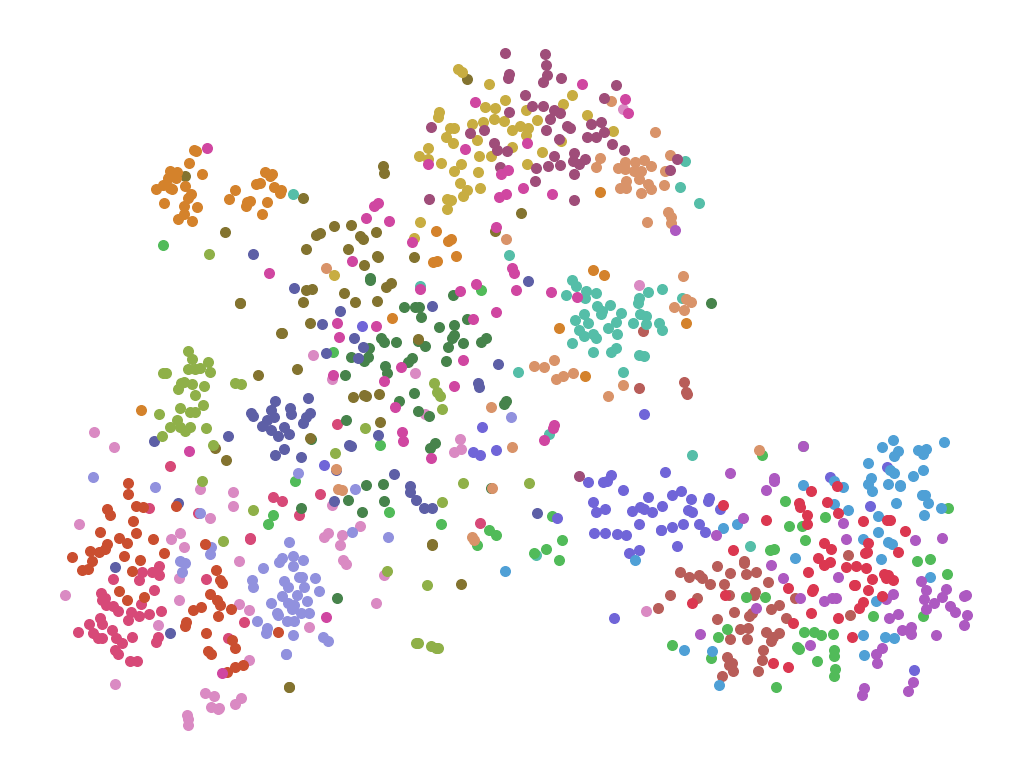}
            \caption{2-dimensional reduction with T-SNE of feature extracted from ImageNet's validation set using a ResNet12 trained on \textit{mini}ImageNet. (Left): images from 20 randomly selected classes represented in \textit{mini}ImageNet's base set.  (Right): Images from the 20 classes represented in \textit{mini}ImageNet's test set. Each color corresponds to a distinct class.}
            \label{fig:clustering-mini-imagenet}
        \end{figure*}

    \magicpar{Measuring the difficulty of outlier detection on novel classes.} \label{subsec:difficulty_of_problem}
        As an anomaly detection problem, open-set recognition consists in detecting samples that differ from the population that is known by the classification model. However, in FSOSR, neither closed-set classes nor open-set classes have been seen during the training of the feature extractor \ie $\mathbb{C}_{base} \cap \cseen =  \mathbb{C}_{base} \cap \cunseen = \emptyset$. In that sense, both the inliers and the outliers of our problem can be considered outliers from the perspective of the feature extractor. Intuitively, this makes it harder to detect open-set instances, since the model doesn't know well the distribution from which they are supposed to diverge. Here we empirically demonstrate and quantify the difficulty of OSR in a setting where closed-set classes have not been represented in the training set. Specifically, we estimate the gap in terms of quality of the classes' definition in the feature space, between classes that were represented during the training of the feature extractor \ie $\mathbb{C}_{base}$, and the classes of the test set, which were not represented in the training set. To do so, we introduce the novel Mean Imposture Factor measure and use the intra-class to inter-class variance ratio $\rho$ as a complementary measure. Note that the following study is performed on whole datasets, \textit{not} few-shot tasks.
            
        \magicpar{Mean Imposture Factor (MIF).}Let $\dphiteta \subset \featurespace \times \mathbb C$ be a labeled dataset of extracted feature vectors, with $\phi_{\thetabf}$ a fixed feature extractor and $\mathbb C$ a finite set of classes. For any feature vector $\zbf$ and a class $k$ to which $\zbf$ does not belong, we define the Imposture Factor $\textit{IF}_{\zbf|k}$ as the proportion of the instances of class $k$ in $\dphiteta$ that are further than $\zbf$ from their class centroid. Then the MIF is the average IF over all instances in $\dphiteta$.

        \begin{align}
            &\boxed{
                \textit{MIF} = \frac{1}{|\mathbb C|} \sum_k \frac{1}{|\dphiteta \backslash \mathcal D_k|} \sum_{\zbf \notin \mathcal D_k} \textit{IF}_{\zbf|k}
            } \\
                ~\textrm{ with  }~
                &\textit{IF}_{\zbf|k} = \frac{1}{|\mathcal D_k|} \sum_{\zbf' \in \mathcal D_k} \mathbbm{1}_{\|\zbf'-\mubf_k\|_2>\|\zbf-\mubf_k\|_2} \nonumber
            \end{align}
        with  $\mathcal D_k$ the set of instances in $\dphiteta$ with label $k$, and $\mathbbm{1}$ the indicator function. The MIF is a measure of how perturbed the clusters corresponding to the ground truth classes are. A MIF of zero means that all instances are closer to their class centroid than any outsider. Note that $\text{MIF} = 1 - \text{AUROC}(\psi)$ where $\text{AUROC}(\psi)$ is the area under the ROC curve for an outlier detector $\psi$ that would assign to each instance an outlier score equal to the distance to the ground truth class centroid. To the best of our knowledge, the MIF is the first tool allowing to measure the class-wise integrity of a projection in the feature space. As a sanity check for MIF, we also report the intra-class to inter-class variance ratio $\rho$, used in previous works \cite{goldblum2020unraveling}, to measure the compactness of a clustering solution.

         \begin{table*}[t]
            \centering
            \small
            \caption{Contrast between datasets made of images from classes represented (\textit{base}) or not represented (\textit{test}) in the feature extractor's training set, on three benchmarks and with several backbones (RN12: ResNet12, WRN: WideResNet1810,  ViT, RN50: ResNet50, and MX: MLP-Mixer), following the MIF (in percents) and the variance ratio ($\rho$). Best result for each column is shown in bold. }
             \resizebox{\textwidth}{!}{
            \begin{tabular}{lcccccccccccccc}
            
            \toprule
             \multirow{3}{*}{Classes} & \multicolumn{4}{c}{miniImageNet} & \multicolumn{4}{c}{tieredImageNet} & \multicolumn{6}{c}{ImageNet $\rightarrow$ Aircraft} \\
                  \cmidrule(rl){2-5} \cmidrule(lr){6-9} \cmidrule(lr){10-15}
                  & \multicolumn{2}{c}{$\rho$} & \multicolumn{2}{c}{MIF (\%)}      & \multicolumn{2}{c}{$\rho$} & \multicolumn{2}{c}{MIF (\%)}      & \multicolumn{3}{c}{$\rho$}            & \multicolumn{3}{c}{MIF (\%)}                      \\
                  \cmidrule(rl){2-3} \cmidrule(lr){4-5} \cmidrule(lr){6-7} \cmidrule(rl){8-9} \cmidrule(lr){10-12} \cmidrule(lr){13-15}
                  & RN12         & WRN         & RN12      & WRN       & RN12         & WRN         & RN12      & WRN       & ViT    & RN50          & MX           & ViT    & RN50          & MX           \\
                  \midrule
            \textit{base} & \textbf{0.93}    & \textbf{0.84}   & \textbf{0.89} & \textbf{1.03} & \textbf{1.09}    & \textbf{0.78}   & \textbf{0.78} & \textbf{0.81} & \textbf{0.96} & \textbf{1.36} & \textbf{2.54} & \textbf{0.09} & \textbf{0.29} & \textbf{0.31} \\
            \textit{test}  & 2.10             & 2.07            & 5.56          & 7.36          & 2.10             & 1.54            & 4.39          & 5.18          & 3.20          & 4.88          & 5.35          & 18.08          & 21.58          & 17.27         \\
            \bottomrule
            \end{tabular}
            }
            \label{tab:clustering-results}

        \end{table*}

        \magicpar{Base classes are better defined than test classes.} We experiment on three widely used Few-Shot Learning benchmarks: \textit{mini}ImageNet \cite{Vinyals16}, \textit{tiered}ImageNet \cite{tiered_imagenet}, and ImageNet $\rightarrow$ Aircraft \cite{maji2013fine}. We use the validation set of ImageNet in order to obtain novel instances for ImageNet, \textit{mini}ImageNet, and \textit{tiered}ImageNet's base classes. We also use it for test classes for consistency.
        In Figure \ref{fig:clustering-mini-imagenet}, we present a visualization of the ability of a ResNet12 trained on \textit{mini}ImageNet to project images of both base and test classes into clusters. While we are able to obtain well-separated clusters for base classes after the 2-dimensional T-SNE reduction, this is clearly not the case for test classes, which are more scattered and overlapping.  Such results are quantitatively corroborated by Table \ref{tab:clustering-results}, which shows that both \text{MIF} and $\rho$ are systematically lower for base classes across 3 benchmarks and 5 feature extractors.
        This demonstrates the difficulty of defining in the feature space the distribution of a class that was not seen during the training of the feature extractor, and therefore the difficulty of defining clear boundaries between inliers and outliers \ie closed-set images and open-set images, all the more when only a few samples are available.

\section{Additional results}

    \begin{figure*}
        \centering
        \includegraphics[width=.99\textwidth]{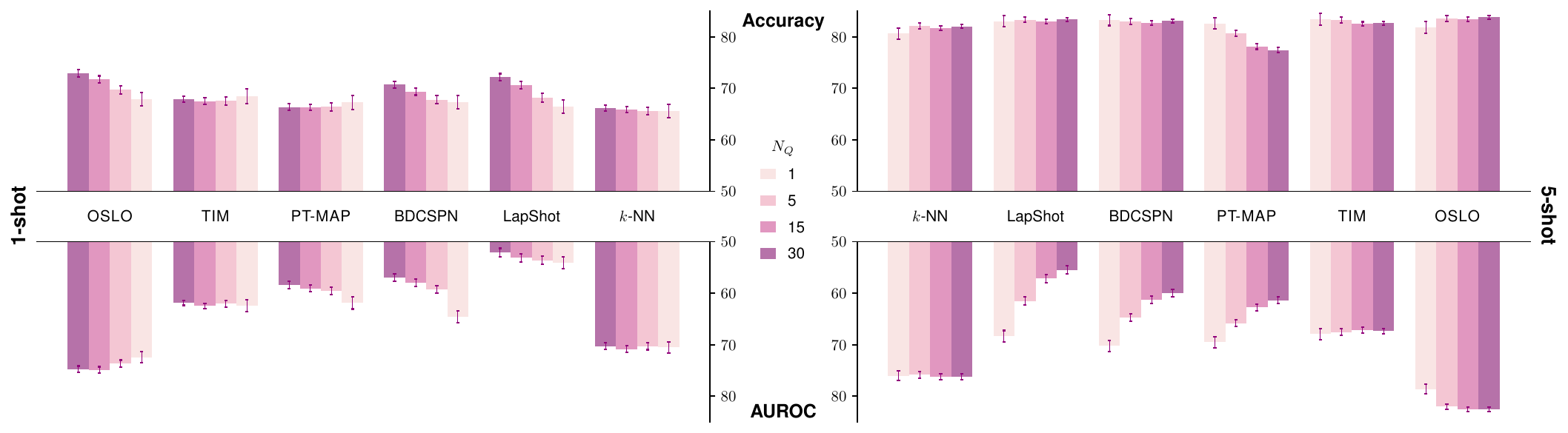}
        \caption{Version of Fig. \ref{fig:variate-query-tiered} on \textit{mini}-ImageNet.}
        \label{fig:variate-query-mini}
    \end{figure*}

\begin{figure*}
        \centering
        \includegraphics[width=0.3\textwidth]{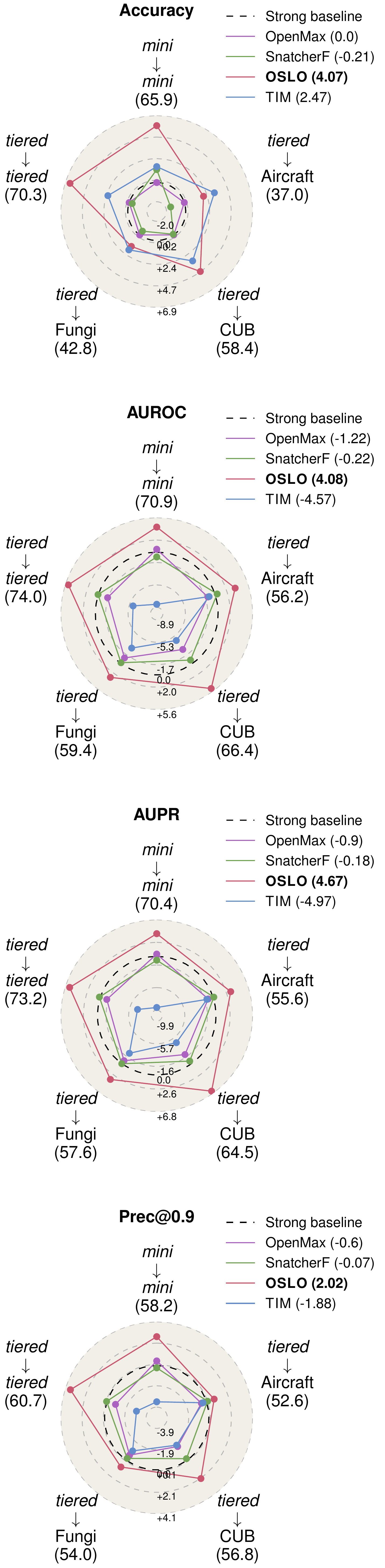}
        \includegraphics[width=0.3\textwidth]{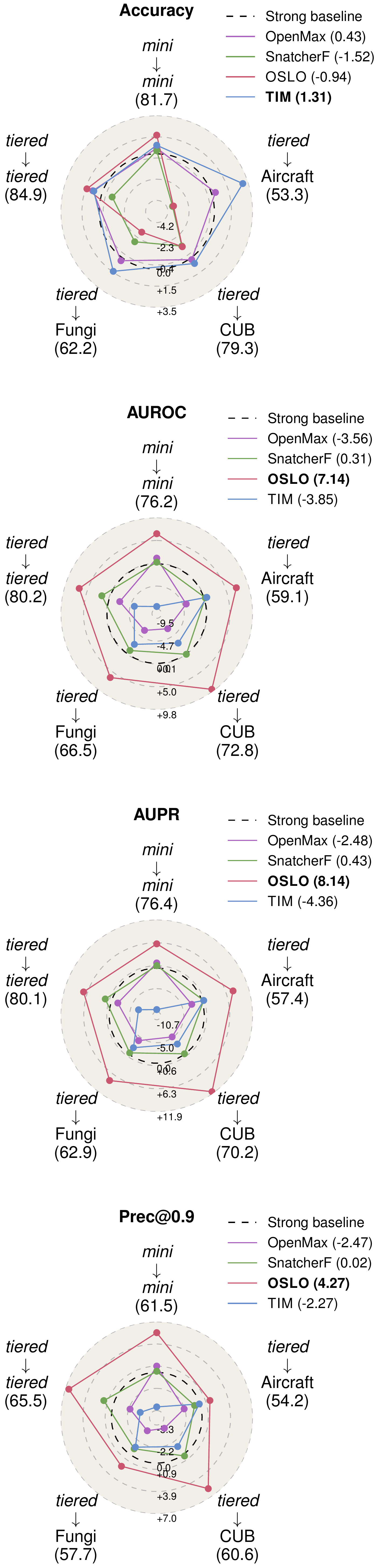}
        \caption{Complete version of Fig. \ref{fig:spider_charts} with a ResNet-12. (Left column): 1-shot. (Right column): 5-shot.}
        \label{fig:full_spider_resnet12}
    \end{figure*}

\begin{figure*}
        \centering
        \includegraphics[width=0.3\textwidth]{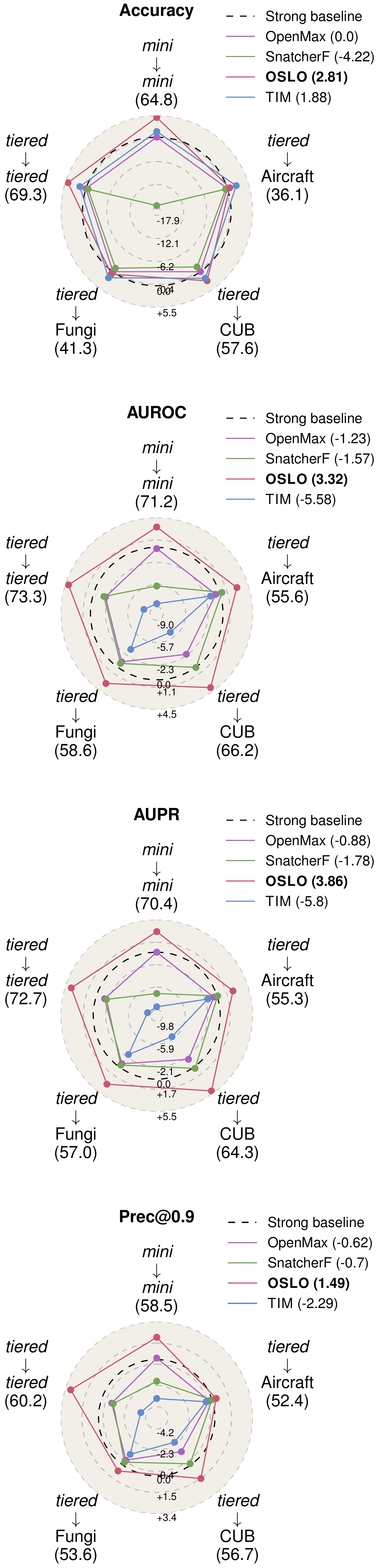}
        \includegraphics[width=0.3\textwidth]{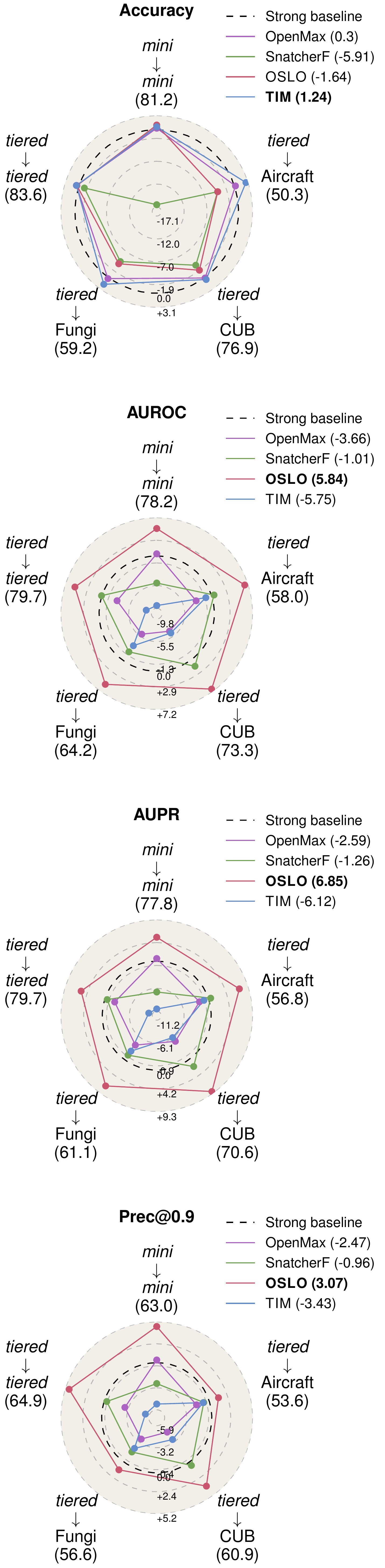}
        \caption{Complete version of Fig. \ref{fig:spider_charts} with a WideResNet 28-10. (Left column): 1-shot. (Right column): 5-shot. SnatcherF was not included in this plot because a yet misdiagnosed problem occurred with the provided \textit{tiered}-ImageNet checkpoint.}
        \label{fig:full_spider_wrn}
    \end{figure*}

In this section we provide more complete versions of plots included in the main paper. Fig. \ref{fig:variate-query-mini} shows the results depending on the size of the query set for \textit{mini}-ImageNet. Furthermore, \ref{fig:full_spider_resnet12} and \ref{fig:full_spider_wrn} complete Fig. \ref{fig:spider_charts}, showing the additional Prec@0.9 metric, along with the results on the WRN2810 provided by \cite{ye2020few}.

\end{document}